\pdfoutput=1
\documentclass[11pt]{article}

\usepackage[]{acl}

\usepackage{times}
\usepackage{caption}
\usepackage{subcaption}
\usepackage{graphicx}
\usepackage{latexsym}
\usepackage{booktabs}
\usepackage{xcolor,colortbl}
\usepackage{amsmath}
\usepackage{multirow}
\usepackage{hyperref}
\usepackage[normalem]{ulem}
\usepackage{lipsum}
\usepackage{footmisc}

\usepackage{amsthm}

\usepackage{tabularx}
\usepackage{cleveref}
\crefformat{footnote}{#2\footnotemark[#1]#3}

\theoremstyle{definition}

\renewcommand{\theexamplex}{%
  \ifnum\value{subsubsection}>0 \thesubsubsection \else
  \ifnum\value{subsection}>0 \thesubsection \else
  \thesection \fi\fi .\arabic{examplex}%
}

\usepackage{amssymb}% http://ctan.org/pkg/amssymb
\usepackage{pifont}% http://ctan.org/pkg/pifont

\usepackage[T1]{fontenc}
\usepackage[utf8]{inputenc}

\usepackage{microtype}
\usepackage{xurl}
\usepackage{pdflscape}
\usepackage{geometry}
\usepackage{longtable}
\usepackage{enumitem}
\usepackage{tipa}

\usepackage{inconsolata}
\usepackage{xcolor}

\definecolor{goldenbrown}{rgb}{0.6, 0.4, 0.08}
\definecolor{jasper}{rgb}{0.84, 0.23, 0.24}
\definecolor{amethyst}{rgb}{0.6, 0.4, 0.8}
\definecolor{airforceblue}{rgb}{0.36, 0.54, 0.66}
\definecolor{navyblue}{rgb}{0.0, 0.0, 0.5}
\definecolor{arylideyellow}{rgb}{0.91, 0.84, 0.42}
\definecolor{kellygreen}{rgb}{0.3, 0.73, 0.09}
\definecolor{orange-red}{rgb}{1.0, 0.27, 0.0}

\newcommand{\cmark}{\textcolor{kellygreen}{\ding{51}}}%
\newcommand{\xmark}{\textcolor{orange-red}{\ding{55}}}%
\def\Plus{\texttt{+}}

\newcommand*{\bigchi}{\mbox{\Large$\chi$}}% big chi

\newcommand{\datasetname}{Belief-R}
\NewDocumentCommand{\normally}{e{^_}}{
  \mspace{3mu} \mid \mspace{-3.6mu} \sim \IfValueT{#1}{#1}
}

\title{Belief Revision: The Adaptability of \\Large Language Models Reasoning}

\author{
    \textbf{Bryan Wilie\thanks{\ These authors contributed equally.}},
    \textbf{Samuel Cahyawijaya$^*$},
    \textbf{Etsuko Ishii},
    \textbf{Junxian He},
    \textbf{Pascale Fung}
    \\
    Hong Kong University of Science and Technology \\
    Clear Water Bay, Hong Kong \\
    \texttt{bwilie@connect.ust.hk} 
}

\begin{document}
\maketitle

% \setcounter{tocdepth}{4}
% \setcounter{secnumdepth}{4}
% \tableofcontents

\begin{abstract}
% claim at the very beginning at the abstract

% The capability to reason from text is crucial for real-world NLP applications, yet most existing evaluations assume that language models (LMs) operate with consistent information. Real-world scenarios, often involve incomplete or evolving data, and in response, individuals update their beliefs and understandings accordingly. 

% The capability to reason from text is crucial for real-world NLP applications. Real-world scenarios often involve incomplete or evolving data. In response, individuals update their beliefs and understandings accordingly. However, most existing evaluations assume that language models (LMs) operate with consistent information. We introduce \datasetname{}, a new dataset designed to test LMs on their belief revision ability when presented with new evidence. This task, inspired by how humans suppress prior inferences, assesses LMs within the newly proposed delta reasoning ($\Delta R$) framework. \datasetname{} features sequences of premises designed to simulate scenarios where additional information could necessitate prior conclusions drawn by LMs. We evaluate $\sim$30 LMs across diverse prompting strategies and found that LMs generally struggle with appropriately revising their beliefs in response to new information. Further, models that adept at updating often underperformed in scenarios without necessary updates, highlighting a critical trade-off. These insights underscores the importance of improving LMs' adaptiveness to changing information, a step toward more reliable AI systems.

The capability to reason from text is crucial for real-world NLP applications. Real-world scenarios often involve incomplete or evolving data. In response, individuals update their beliefs and understandings accordingly. However, most existing evaluations assume that language models (LMs) operate with consistent information. We introduce \datasetname{}\footnote{\ The code and dataset are available at \url{https://github.com/HLTCHKUST/belief-revision}}, a new dataset designed to test LMs' belief revision ability when presented with new evidence. Inspired by how humans suppress prior inferences, this task assesses LMs within the newly proposed delta reasoning ($\Delta R$) framework. \datasetname{} features sequences of premises designed to simulate scenarios where additional information could necessitate prior conclusions drawn by LMs. We evaluate $\sim$30 LMs across diverse prompting strategies and found that LMs generally struggle to appropriately revise their beliefs in response to new information. Further, models adept at updating often underperformed in scenarios without necessary updates, highlighting a critical trade-off. These insights underscore the importance of improving LMs' adaptiveness to changing information, a step toward more reliable AI systems.

% necessitating a dynamic update of beliefs in light of new information.
\end{abstract}

\begin{figure}
    \centering
    \resizebox{\linewidth}{!}{
    \includegraphics{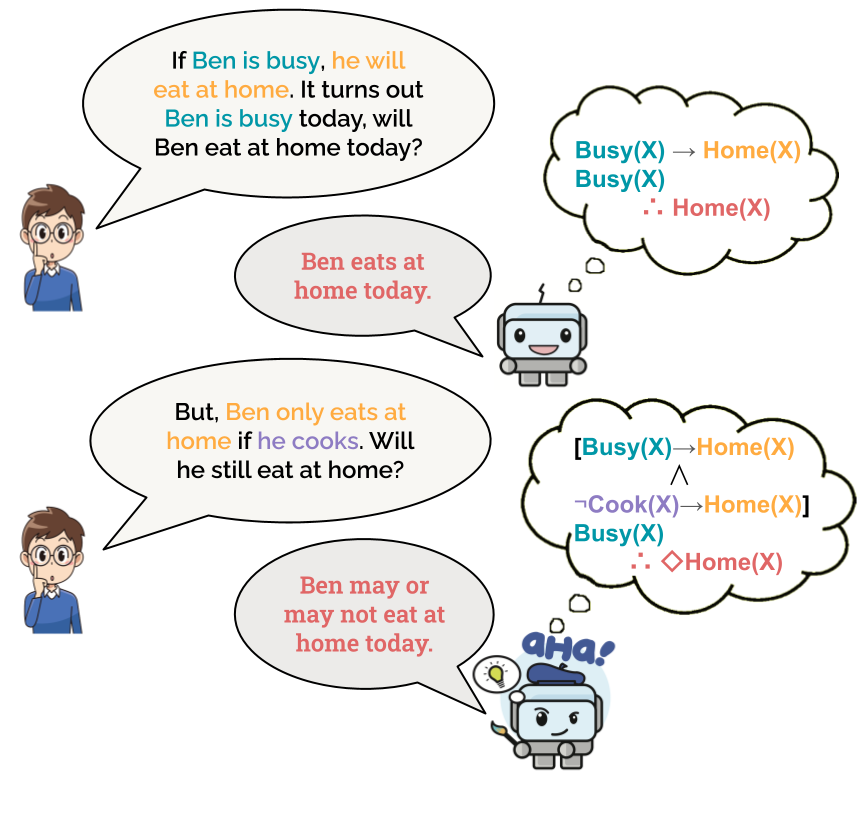}
    }
    \vspace{-20pt}
    \caption{Belief revision allows reasoners to update their belief based on the new provided evidence. Such ability is necessary to enable better logical reasoning on the case of defeasible inference.}
    \label{fig:enter-label}
    \vspace{-8pt}
\end{figure}

\section{Introduction}

% “[...] if a machine is expected to be infallible, it cannot also be intelligent.”
% - Turing
% Argumentation: A calculus for Human-Centric AI

% Reasoning is a cognitive process that involves applying evidence, arguments, and logic to reach conclusions or make decisions. 
% Human reasoning is characterized by its ability to deal with incomplete information. We draw conclusions with the available evidences, but in considerations on further information, our conclusions might be invalidated. In this situation, we give up the previous ones and derive new alternative conclusions instead~\cite{lukaszewicz1990non, brewka1991nonmonotonic}.

Human reasoning is characterized by its ability to deal with partial or evolving information. When new information becomes available, we dynamically update our beliefs. We reevaluate and adjust our initial premises or conclusions as necessary in light of this new evidence~\cite{lukaszewicz1990non, brewka1991nonmonotonic}. For instance, knowing \textit{Tweety is a bird}, we conclude that \textit{it flies} since \textit{birds usually fly}. Discovering \textit{Tweety is a penguin}, we retract the conclusion but not the other premises; we still believe \textit{Tweety is a bird} and that \textit{birds typically fly}, however, we now conclude that \textit{it cannot fly} since we know that \textit{penguins cannot fly}.
This form of reasoning permits new information to undermine prior beliefs, which necessitates the ability of \textit{belief revision}~\cite{gardenfors1988knowledge, gardenfors1991belief,rott2001change}.
% wobcke1995belief
% bochman2001logical
% This type of reasoning is known as \textit{nonmonotonic reasoning}~\cite{reiter1988nonmonotonic, antoniou1997nonmonotonic}.
% huang2023large
% brewka2008nonmonotonic

% exhibiting a kind of nonmonotonic reasoning~\cite{mcdermott1980non, reiter1988nonmonotonic}, 
% Such form of reasoning, which allow additional information to invalidate previous conclusions, are called \textit{nonmonotonic reasoning}~\cite{reiter1988nonmonotonic, antoniou1997nonmonotonic}.
% Non-monotonic reasoning has been of little attention in modern NLP research~\cite{pfeifer2005coherence, brahman2021learning}. 

% This characteristic aligns with cognitive science's concept of "belief revision," where humans revise their beliefs to maintain coherence with newly integrated data.
% real-world problems~\cite{kraus1990nonmonotonic, strasser2018non}, 
% everyday reasoning~\cite{strasser2018non}, 

\begin{table*}
\centering
\resizebox{\textwidth}{!}{%
\begin{tabular}{lccccccccc}
\toprule
\textbf{Features} & \textbf{bAbI 15} & \textbf{FOLIO} & \textbf{Proof Writer} & \textbf{Leap of Thought} & \textbf{$\alpha$NLI} & \textbf{BoardgameQA} & \textbf{PropInd} & \textbf{\datasetname{}} \\
\midrule
Incomplete info & \xmark & \xmark & \xmark & \cmark & \cmark & \cmark & \cmark & \cmark \\
Contradictory info & \xmark & \xmark & \xmark & \xmark & \cmark & \cmark & \cmark & \cmark \\
Belief revision & \xmark & \xmark & \xmark & \xmark & \xmark & \xmark & \xmark & \cmark \\
\bottomrule
\end{tabular}%
}
\caption{
The comparison of \datasetname{} with other widely-used logical reasoning datasets.
\datasetname{} uniquely examines scenarios potentially necessitating belief updates.
% in response to new information. 
\datasetname{} specifically evaluates the capability of belief revision, assessing whether prior beliefs should be adjusted or retained depending of the significance of the new information.
% \jh{It is visually better to use red and green colors for xmark and checkmark respectively.}
}
\label{tab:dataset_comparison}
\vspace{-10pt}
\end{table*}

The ability to adjust beliefs allows better adaptability of AI systems by enabling them to properly revise prior inferences
% drawn from changing information 
as further evidence emerges, such as in commonsense inferences~\cite{brewka1997nonmonotonic, etherington1986reasoning, pfeifer2005coherence} and decision-making~\cite{antoniou1997nonmonotonic, dubois2002qualitative}. Despite this, recent reasoning evaluations of state-of-the-art AI technologies, such as language models (LMs), primarily focus on its ability to draw conclusions assuming complete information (c.f.~\citet{bhagavatula2020abductive, han2024inductive, kazemi2024boardgameqa}). While these evaluations useful to demonstrate the reasoning abilities of LMs, they fail to capture the concept of belief change.

We introduce \datasetname{}, the first-of-a-kind diagnostic reasoning evaluation dataset designed to assess inferences involving belief revision. \datasetname{} is inspired by the concept of the Suppression Task~\cite{byrne1989suppressing} which enables the retraction of previously inferred beliefs by the introduction of new contextual premises, mimicking how humans reassess their inferences when presented with additional context. To allow a specific and measurable evaluation on belief revision, we introduce a new reasoning evaluation setting dubbed as \textbf{delta reasoning ($\mathbf{\Delta R}$)} framework. Within $\mathbf{\Delta R}$, evaluation is done within two sequential reasoning steps. We start by presenting LMs with two initial premises that satisfy basic logical inference rule to assess its basic inference ability. We expect the model to make accurate inferences to establish the prior beliefs. Then, we introduce another premise to see if the model adjusts its beliefs or keeps them unchanged, depending on the significance of the newly introduced information to the initial beliefs.

\datasetname{} is specifically designed to support the belief revision evaluation through the $\mathbf{\Delta R}$ framework. Each sample in \datasetname{} is equipped with two initial premises that support basic modus ponens or modus tollens inferences, and a new premise that brings in new information that might modify previously held beliefs. We synthetically generate the premises in \datasetname{} leveraging on publicly available dataset, and manually annotate the new information significances along with the ground truth answers through multiple human annotators and majority voting. As illustrated in Figure~\ref{fig:nmr_illustration}, \datasetname{} uniquely facilitates thorough evaluations of belief revision capabilities.

Through \datasetname{}, we evaluate the belief revision ability of small and large scale LMs using different prompting techniques. Our study shows that these models are incapable of revising their prior beliefs. We further reveal a critical limitation: they confront a performance trade-off between updating and maintaining their prior beliefs. Models that perform better in the cases where an update is needed, typically faltered on the other. Furthermore, better prompting methods also fail to significantly enhance this capability. These insights underscore a need for strategies to enhance model's capability to correctly update or maintain its initial beliefs when faced with new evidence to ensure its reliability across evolving scenarios.

% This issue is even more pronounced in smaller models which have a hard time forming initial beliefs.
% In our experiments, we found that it's difficult to establish prior beliefs in smaller-scale LMs. 

% We benchmark various LMs on BoardgameQA and measure their defeasible reasoning capacity. Most notably, our results reveal that LMs perform poorly when reasoning with conflicting sources, especially in the few-shot setting (compared to the finetuning setting) suggesting that preference understanding and defeasible reasoning capacities do not surface out-of-the-box in pretrained LMs. Secondly, we find that smaller LMs perform poorly when not all of the required information is provided as input. These results highlight a critical gap in the reasoning capacity of current LMs, considering that reasoning over contradicting and incomplete sets of information is a common scenario in many applications, and is key for developing robust AI systems.

\section{Related Works}
\label{sec:related_works}
\paragraph{Belief revision}

Belief revision is the process of changing beliefs to take into account a new piece of information. In AI systems, one of its early implementation is through procedures by which databases can be updated, i.e. for recording and maintaining reasons for system beliefs~\cite{doyle1979truth, falappa2002explanations, sep-logic-belief-revision}. Notably,~\citet{alchourron1985logic} created formal frameworks to determine how beliefs should be updated in a rational manner. The core challenge in belief revision is deciding rationally which prior beliefs to modify, retain, or discard when confronted with new evidence~\cite{rott2001change}. Consequently in this paper, we look at how LMs handle belief revision. Belief in LMs can be thought of as models' output
% or what it considers to be true
~\cite{li-etal-2019-logic, jang2022becel, wang2023can}.
Several works revise LMs' beliefs through updating its parameter directly or via fine-tuning~\cite{de2021editing, dai2021knowledge, hase2023methods}. However, this process is not a rational process of the model itself~\cite{hofweber2024language}. Moreover, it relies on pre-prepared knowledge, which is not ideal if we envision LMs to help with discovering new things~\cite{ban2023query, ma2024llm}. In this work, we assess LMs' belief revision capabilities through its response towards queries that neccessitate judgement on whether it needs to update its prior beliefs or keep it.

% While databases are persistent and be viewed as belief, one of the main challenge in LLM is that they can still produce inconsistent answers when probed, and thus its belief is not apparent~\cite{kassner2021beliefbank, elazar2021measuring, jang2022becel, hase2023methods}. ~\cite{wang2023can} found that whether LMs can maintain their beliefs in truth even when challenged by oftentimes absurdly invalid arguments.
% In our experiments, we ensured prior belief through basic logical inference.

\paragraph{Language model reasoning evaluation} 

Reasoning is one of the fundamental intelligent behaviors, essential for solving complex real-world tasks~\cite{huang-chang-2023-towards}. One works test this behaviour by creating simple tasks to comprehensively check if a system can answer questions by connecting facts or using basic logic~\cite{weston2016towards}. Others design more advanced tests to evaluate inductive, deductive, and abductive reasoning~\cite{sinha2019clutrr, saparov2024testing, bhagavatula2020abductive}. Some benchmarks replicate real-world complexities by presenting partial or conflicting informations~\cite{arabshahi2021conversational, sprague2022natural, han2024inductive, kazemi2024boardgameqa}. Belief revision focuses on the adaptability problem: whether the model properly revises prior beliefs as new information emerges. We extend the reasoning evaluation by focusing on scenarios where information evolves, presenting queries that require dynamic updates of prior beliefs in light of new evidence. This is distinct from other existing reasoning tasks involving incomplete or contradictory information, i.e. in~\citet{talmor2020leap, bhagavatula2020abductive, kazemi2024boardgameqa}, since they assume a static environment and focus on filling the gap or resolving the contradiction. We further note the comparison in Table \ref{tab:dataset_comparison}.

% % Despite its fundamental role in human reasoning
% While humans excel at this type of reasoning, 
% % creating AI systems capable of non-monotonic reasoning across a broad spectrum of situations expressed in natural language remains a significant, open research challenge
% even the most advanced AI systems, i.e., large language models (LLMs), largely fail on non-monotonic reasoning across a broad spectrum of situations expressed in natural language
% % Current works on large language models (LLMs) have demonstrated significant progress in various reasoning benchmarks
% ~\cite{bang2023multitask, touvron2023Llama, openai2023gpt4}.
% However, existing assessments of reasoning capabilties often overlook the complexities of non-monotonic contextual elements (c.f.~\citet{weston2016towards, sinha2019clutrr, creswell2022selection}). 
% It is also observed that LLMs exhibit limitation in handling tasks that requires nonmonotonic reasoning. These limitations include challenges in abductive reasoning, in tasks involving belief revision, conflict resolution, and in knowledge or property inferences~\cite{liu-etal-2023-magic, wang2023can, han2024inductive}.
% Furthermore, current assessments of reasoning in LMs encompass a broad range of formal logic and natural language interpretations~\cite{weston2016towards, sinha2019clutrr, bhagavatula2020abductive, wan2024b} but often overlook the complexities of non-monotonic contextual elements.
% , 
% They all share a fundamental issue: the need to assess implications in a non-monotonic manner. 

\section{Belief Revision}

% In reasoning, our main goal is to reach true conclusions based on supportive evidence.
% Belief revision is the ability to adapt and evolve the reasoning process based on new evidences, leading the reasoner to adjust and modify its prior beliefs to reach new conclusions. Possessing the ability to perform belief revision is critical as it ensures rational decision-making despite the incomplete and evolving nature of available information over time~\cite{brewka1997nonmonotonic, nute2001defeasible, makinson2005relations, ribeiro2019belief}. 

% Belief in LMs' can be thought of as models' output or what it considers to be true~\cite{li-etal-2019-logic, jang2022becel, wang2023can}. Consequently, 
% \jh{Generally in this section, I think we should explicitly define ``modus ponens'' and ``modus tollens'' in a clear way because later in the paper you just write these assuming readers know. For many readers like me, I do not have a clear notion on what are ``modus ponens'' and ``modus tollens''.}
Belief revision is the ability to adapt the reasoning process in response to new information.
% thus to revising its prior beliefs to form new output.
This capability is critical as it ensures rational decision-making in the face of incomplete and evolving nature of available information~\cite{nute2001defeasible, makinson2005relations, ribeiro2019belief}. 
% In this section, we elaborate the concept of belief revision and its notation, and introduce our proposed framework to evaluate the belief revision capabilities.
In this section, we introduce the concept of belief revision and its notation, and propose the evaluation framework for belief revision capabilities.

% brewka1997nonmonotonic

% \cite{jang2022becel}
% Belief -> what a model considers to be true
% In this regard, a model with a high level of NLU ability should capture the meaning in essence and make the same decisions in semantically identical texts considering the definition of “understanding language”, and this is the concept of semantic consistency. The belief and principle become a model’s predictions on semantically identical texts and semantic equivalence, respectively. So, we define the semantic consistency of an LM as its ability to make the same decisions on semantically equivalent texts
% the belief and principle become a model’s predictions regarding instances where the logical property holds and a logical property.
% The basic concept of factual consistency is that a model should generate factually accurate outputs. Therefore, the belief is the model’s output

% \cite{li-etal-2019-logic}
% Reasoning about language requires that a
% system has the ability not only to draw correct inferences about textual inputs, but also to be consistent its beliefs across various inputs.

\subsection{Background and notation}

% \cite{feng2023towards, van-niekerk-etal-2020-knowing}

% We let $\Gamma{=}\{\gamma_1,\dots,\gamma_N\}$ represent a set of premise sentences that could imply conclusion sentences $\Phi{=}\{\varphi_1,\dots,\varphi_M\}$. 
For set of query sentences $\bigchi$, it encompasses a set of premises $\Gamma{=}\{\gamma_1,\dots,\gamma_N\}$ that could imply a set of conclusions $\Phi{=}\{\varphi_1,\dots,\varphi_M\}$. In this work, we conceptualize "belief" similarly to its usage in dialogue systems, where it represents what the system currently considers true based on the context~\cite{feng2023towards, van2020knowing}.
We denote reasoner's belief set as a set of sentences $\mathcal{B}$ to represent a contextually fixed background knowledge of $\bigchi$. In this regard, $\mathcal{B}$ is a tuple that contains set of premises and conclusions: $\mathcal{B}{=}(\Gamma,\Phi)$. In presence of new information $\gamma_{N\Plus1}$, the belief revision concept allow us to infer conclusion $\varphi_{M\Plus1}$ if it is rational to believe $\varphi_{M\Plus1}$ after acknowledging $\gamma_{N\Plus1}$.

\paragraph{Belief revision operation} The belief revision operation is to update belief set \( \mathcal{B} \) with a new piece of information, \( \gamma_{N\Plus1} \). Here, the result of operation must always be that the beliefs does not contradict one another to avoid inconsistencies among them. The significance of the new information \( \gamma_{N\Plus1} \), decides whether it fits with or modifies the existing beliefs after performing the belief revision operation. The operation should smoothly incorporate \( \gamma_{N\Plus1} \) and yield a new conclusion \( \varphi_{M\Plus1} \) as long as it does not conflict, thereby justifying the maintenance of the reasoner’s prior beliefs. However, if it conflicts, we update the initial beliefs $\mathcal{B}$ appropriately, i.e., by retracting any prior conclusions in $\Phi$, to incorporate the new, conflicting information \( \gamma_{N\Plus1} \) to resolve any inconsistencies as we yield the correct $\varphi_{M\Plus1}$.
The process to figure out what follows from the revised beliefs is then essentially to infer the new conclusion $\varphi_{M\Plus1}$.

\begin{table*}[!t]
\small
\centering
\resizebox{0.95\linewidth}{!}{%
\begin{tabularx}{\columnwidth}{p{7cm}}
\toprule
If she has an essay to finish then she will study late in the library \\
She has an essay to finish \\
\textbf{\textcolor{amethyst}{If the library stays open then she will study late in the library}}\\ \\
What necessarily had to follow assuming that the above premises were true? \\
(a) She will study late in the library. \\
(b) She will not study late in the library. \\
(c) She may or may not study late in the library. \checkmark  \\
\bottomrule
\end{tabularx}%
\begin{tabularx}{\columnwidth}{p{7cm}}
\toprule
If she has an essay to finish then she will study late in the library \\
She has an essay to finish \\
\textbf{\textcolor{airforceblue}{If she has some textbooks to read then she will study late in the library}}\\ \\
What necessarily had to follow assuming that the above premises were true? \\
(a) She will study late in the library. \checkmark \\
(b) She will not study late in the library. \\
(c) She may or may not study late in the library.\\
\bottomrule
\end{tabularx}%
}
\captionof{figure}{
% [INSERT ABDUCTIVE REASONING TASK HERE].\\
% Non-monotonic reasoning involves forming and revising plausible conclusions based on evolving evidence. 
Human reasoning adapts based on new information, leading us to adjust our prior beliefs.
% The statement in \textbf{bold} in the pragmatic reasoning example invalidates the previous assumptions, i.e.\ "\textit{it's common for an employee to be nervous when entering their boss's office}". 
% In this example of suppression task~\cite{byrne1989suppressing},
Here, \textbf{\textcolor{amethyst}{the additional condition (left)}} casts doubt on prior modus ponens conclusion in (a).
% "\textit{she will study late in the library}"
People may consider that certain other conditions necessary for this conclusion to hold, i.e., \textit{the library must remain open}. In contrast, \textbf{\textcolor{airforceblue}{the alternative argument (right)}} does not affect the modus ponens inference pathway, thus prior conclusion could still hold.
% \jh{Maybe make it more explicit in the table on ``additional condition'' and ``alternative argument''. Currently colors are the only signal to relate to the captions, but keep in mind that some people may not read the paper in color (like bw printing)}
% is not a necessary conditions for revising the prior belief. 
% Even if \textit{she doesn't have any textbooks to read}, \textit{she has an essay to finish anyway}, so \textit{she will still study late in the library}.
% 
}
% Example of suppression task~\cite{byrne1989suppressing}. In the first case, the premises support the conclusion: \textit{She will study late in the library}. However, when the premises are accompanied by another additional conditional (in \textbf{bold}), most of us tend to reject the previously drawn conclusion, since the possibility that \textit{the library does not stay open} casts doubt on this conclusion.}
\label{fig:nmr_illustration}
\vspace{-8pt}
\end{table*}

\subsection{Evaluating belief revision with $\mathbf{\Delta R}$}

% On a given condition of the input query $\mathcal{Q}$, an LLM reasoner $\mathcal{M}$ auto-regressively generates its responses $\mathcal{R}$ either directly or through chain-of-thought reasoning steps.
% Let $\bigchi$ be the query we feed to the model for its response $\mathcal{A}$.
% \( \alpha_{t\Plus1}{=}\{\varphi_{M\Plus1}\}\), 

We introduce a novel \textbf{delta reasoning ($\mathbf{\Delta R}$)} framework, to study how LMs adapt their reasoning when presented with new information over successive timesteps. In this framework, we focus on understanding how model responds to query changes at two essential, consecutive reasoning steps at \( t \) and \( t\Plus1 \).
We do this by comparing responses to prior queries at step $t$, \(\chi_t\), and the next query at step $t\Plus1$, \(\chi_{t\Plus1}\), adding the new information \( \gamma_{N\Plus1}\).

To begin with, we need $\chi_t$ to minimally include two premises, i.e.\ $\{\gamma_1,\gamma_2\}$, and at least imply conclusion $\varphi_1$. We set $\chi_t$ to be basic as we expect LMs to answer it in high accuracy to help establish the prior belief and not be affected by the inconsistencies in LMs' behaviour~\cite{jang2022becel, kassner2021beliefbank, hase2023methods}. We then add the new information \( \gamma_{N\Plus1}\) as another premise $\gamma_3$ in $\chi_{t\Plus1}$ such that $\chi_{t\Plus1}{=}\{\gamma_1,\gamma_2,\gamma_3\}$. We examine the corresponding conclusion, $\varphi_{M\Plus1}$, to see how the beliefs shifts according to the significance of \( \gamma_3\).

One way to set $\chi_t$ as basic, is to state them as premises that could satisfy basic logical inference rules of modus ponens and modus tollens~\cite{wason1972psychology, haack1978philosophy, evans1982psychology}. Modus ponens and modus tollens is a valid form of inference that have been made a central principle in many propositional and modern logics~\cite{copi1972introduction, haack1978philosophy}. Modus ponens rule of inference states that the premises ``if $p$ then $q$'' is true and $p$ is true ($p\rightarrow q, p$) satisfy modus ponens conclusion that $q$ must be true ($q$). Modus tollens rule of inference states that the premises ``if $p$ then $q$'' is true and $q$ is false ($p\rightarrow q, \neg q$) satisfy modus tollens conclusion that $p$ must be false ($\neg p$).

In this setup, we are able to evaluate how well the models revise its beliefs after the introduction of new information in \( \gamma_3 \). We measure the model's dynamic reasoning ability: whether it can correctly update or maintain its initial beliefs when confronted with new information that may contradict prior beliefs. Through this approach, we can assess both how accurate and how flexible different reasoning models are in evolving scenarios.

% We refer to these model’s generations $\mathcal{R}$ as “beliefs” following~\citet{kassner2023language, wang2023can} as any agent can be said to believe $p$ if it acts as if $p$ was true ~\cite{Schwitzgebel_2019}.
% % We limit our elaboration on "LLM belief", since currently it is not always clear how or if an answer follows from their latent “beliefs” about the world, or whether the LLM even has a coherent internal belief system.

% as control in the comparison with $\mathcal{R}_{t+1}$.
% as the belief at step $t$ and $t+1$.

% We set $\mathcal{Q}_t$ to be basic as we expect the model to answer in high accuracy to help establish the prior belief.
% Afterwards, at step $t\Plus1$, when we ask the model to respond the query $Q_{t\Plus1}$, we expect it to judge whether it needs to update its previous conclusions at step $t$ or maintain them, depending on the significance of the new information \( p_{N\Plus1}\).
% % , based on the prevalence $r$ in $x$ relative to the initial premises at $\mathcal{Q}_t$. 
% In this setting, we can evaluate whether the model produces $\mathcal{R}_{t\Plus1}$ accordingly in both cases: where the ground-truth is for it to be revised or not. 
% % depending on the prevalence $r$ is strong or weak.

% \begin{example}
\paragraph{Example}\ Figure \ref{fig:nmr_illustration} presents a scenario where the initial two premises at step $t$, $\gamma_1$ and $\gamma_2$, adhere to a basic inference rule, modus ponens ($p\rightarrow q,p\vdash q$), implying a $\varphi_1$ conclusion of $q$:
% \jh{what is $q$? It seems we do not define q in the paper?}
\textit{She will study late in the library}. These premises: $\gamma_1$, $\gamma_2$, and $\varphi_1$, form the belief set $\mathcal{B}$. Subsequently, we introduce the third premise $\gamma_3$, i.e., another conditional ($r\rightarrow q$)
% \textbf{\textcolor{amethyst}{purple}}
``\textcolor{amethyst}{if the library stays open then she will study late in
the library}'',
% or \textbf{\textcolor{airforceblue}{blue}}) 
as the new information in query $\chi_{t\Plus1}$ and evaluate model's answer at step $t\Plus1$. 
This sets the stage to execute the belief revision operation.
% This sets the stage to execute the inference operation $\gamma_3 {\normally}_\mathcal{B} \varphi_2$ as follows:

\textit{Recall $\mathcal{B}$ = \{$\gamma_1$ : \textit{If she has an essay to finish then she will study late in the library}., $\gamma_2$ : \textit{She has an essay to finish.}, $\varphi_1$ : \textit{She will study late in the library.}\}, and $\gamma_3$ = \{\textit{If the library stays open then she will study late in the library.}\}. 
% The introduction of new condition $\gamma_3$ indicates that the prior conclusion depends on whether the library is open or not.
% \jh{As I already discussed with Bryan on this, $\gamma_3$ here only literally states ``library is open'' is an sufficient but unnecessary condition. I don't think it directly reflects ``the prior conclusion depends on whether the library is open or not'' (which requires it to be a necessary condition), a commonsense reasoning step is involved to say this.} 
% This raises doubt on the validity of the formerly accepted conclusion $\varphi_1$,
The introduction of $\gamma_3$ suggests that ``the library being open'' is a sufficient condition for her to ``study late in the library''. However, people might consider it as a necessary condition for $\varphi_1$. This would involve commonsense reasoning step to recognize that despite the conditions set by $\gamma_1$ and $\gamma_2$, the actual feasibility of her studying late as concluded in $\varphi_1$ might inherently depend on the library's availability.
% However, to consider it a necessary condition for $\varphi_1$, a commonsense reasoning step is essential. This would involve recognizing that despite the conditions set by $\gamma_1$ and $\gamma_2$, the actual feasibility of her studying late as concluded in $\varphi_1$ might inherently depend on the library's availability. 
Thus, while $\gamma_3$ does not explicitly redefine the dependency of $\varphi_1$ on the library's status, it implies a scenario where such a dependency could be reasonably inferred.
Consequently, we retract $\varphi_1$ and infer the new conclusion $\varphi_2$: ``She may or may not study late in the library''.}
% $\mathcal{P}$ = \{$d_1$ $\prec$ $f_3$\}. 
% \textit{Here, the nonmonotonic theory deals with the conflicts in the derived valid inference $d_1$ and the latter introduced conditional $f_3$. Since it's implied that $d_1$ $\prec$ $f_3$, we should retract conclusions derived from $d_1$, if any, since we first need to know whether the library will remain open or not.}
% \end{example}

% \section{Nonmonotonic Reasoning Capabilities of Large Language Models}
\vspace{-4pt}

\section{The \datasetname{} Dataset}

% We investigate the conclusions derived through logical inference using the suppression task~\cite{byrne1989suppressing} 
% and through contextual understanding in pragmatic reasoning on implied sarcasm}.

% \paragraph{Suppression of valid inferences}
% We design \datasetname{} dataset to aim at assessing the capacity for nonmonotonic reasoning in language models, specifically in scenarios involving belief revision within the $\mathbf{\Delta R}$ framework.
\datasetname{} is designed to specifically assess the belief revision capability through the $\mathbf{\Delta R}$ framework.
To account for this, we adopt a reasoning task that has been extensively studied in cognitive science: the suppression task~\cite{byrne1989suppressing}. Typically, this task employs a trio of premises $\gamma_1,\gamma_2,\gamma_3$ that accompanied by three possible conclusions, i.e.\ as exemplified in Figure~\ref{fig:nmr_illustration} for modus ponens: (a) \textit{She will study late in the library} ($q$), (b) \textit{She will not study late in the library} ($\neg q$), and (c) \textit{She may or may not study late in the library}
% ($\Diamond q \land \Diamond \neg q$).
($\Diamond q \land \Diamond \neg q$; here the symbol $\Diamond$ expresses possibility, $\Diamond q$ can be read as ``possibly $q$'').

At step $t$, we form a query $\chi_t$ using the first two premises, $\gamma_1$ and $\gamma_2$. These two premises are the premises that respectively satisfy the modus ponens or modus tollens conclusion, ($p\rightarrow q, p$) or ($p\rightarrow q, \neg q$).
% \jh{I don't understand why the two premises can be written like this, plz make it more clear}
% and they respectively satisfy the conclusion modus ponens or modus tollens. 
These logical rules are basic, and we generally expect that most reasoners can apply them accurately. Next, at step $t\Plus1$, to form the query $\chi_{t\Plus1}$, we introduce a third premise $\gamma_3$ which is another conditional statement $r \rightarrow q$. The addition of $\gamma_3$ brings in new information that might conflict previously held beliefs. The new information in $r$ can be seen either as adding more requirements or providing an alternative pathway, i.e.\ to reach the same modus ponens conclusion $q$.

For instance, if $\gamma_3$ states \textit{if the library stays open then she will study late}, we now view $r$: \textit{the library stays open} as another \textbf{\textcolor{amethyst}{additional}} requirement on top of $p$. In such cases, just knowing $p$ alone isn't enough to conclude $q$:
% \jh{The notations are really confusing to me in this paper. I feel we kinda have two notation systems in the paper -- when defining the problems we use $\gamma_1, \gamma_2, \gamma_3$ stuff to denote premises and $\phi$ to denote the conclusion as in Section 3.1, which is clear. However, when we introduce more we gradually use more of $p$, $q$, and here $r$ sometimes, and $p$ and $q$ also denote premises and conclusions? If it is not necessary to have so many different notations, plz try to use fewer notation symbols for easier understanding. Also, if we use $p, q, r$ so frequently, then they should be defined in Section 3.1 as well.}
we also need $r$ to be true, thus the condition now becomes $p\land r {\rightarrow} q$. In this case, we retract the prior modus ponens conclusion $q$, and infer the new conclusion $\Diamond q \land \Diamond \neg q$. We refer to this subset of dataset as the \textbf{\textcolor{amethyst}{``Belief Update'' (BU)}} category. However, in another case, $\gamma_3$ could instead states \textit{if she has textbooks then she will study late}. In this case, $r$ stands as a separate \textbf{\textcolor{airforceblue}{alternative}} inference path that also leads to $q$, thus $p\lor r {\rightarrow} q$. Here, $p$ still directly leads to $q$, and the acknowledgement of $r$ doesn't affect this pathway, enabling prior conclusion to still hold. We call this subset as the \textbf{\textcolor{airforceblue}{``Belief Maintain'' (BM)}} category.

% \datasetname{} task is for the reasoner to handle contextual relevancy of the information in $r$ and to judge whether it needs to update its prior beliefs at step $t$. To do so, the model needs to reason about the commonsense and causal dependencies between the provided three premises, to determine the relationship between the $p$ and $r$ in the premises, whether it's an $\loand$ or $\lor$. 

% In \datasetname{}, the task requires the model to manage and evaluate the relevance of information within $r$ across multiple steps, deciding whether to update its beliefs at each step $t$.

In \datasetname{}, the task requires the model to perform multi-step reasoning to manage the relevance of information within $r$ and decide if it needs to update its prior beliefs at step $t$ or not. The model must discern the implicit commonsense and causal links amongst given premises to identify how $p$ and $r$ are related, determining if their interaction is conjunctive ($p\land r$) or disjunctive ($p\lor r$). Based on the relationships, reasoner needs to determine whether to update its initial conclusion $q$ if the new information $r$ imply an additional requirement for its prior beliefs to hold ($p\land r$), or to maintain its prior beliefs if $r$ simply serves as alternatives ($p\lor r$). To quantitatively measure the model's reasoning accuracy, we provide multiple choices and ask it to pick the most plausible conclusion. For instance, in examples shown in Figure~\ref{fig:nmr_illustration}, we would expect LMs to choose options (c) and (a) for each scenario, which aligns with the majority choices made in the original study \cite{byrne1989suppressing,byrne1999counterexamples}.
% according the majority votes in the original test as their final answer~\cite{byrne1989suppressing,byrne1999counterexamples}.

\subsection{Dataset construction} 
We leverage ATOMIC~\cite{sap2019atomic}, a publicly-available dataset of everyday commonsense reasoning. It contains textual descriptions of inferential if-then knowledge (e.g., “if X pays Y a compliment, then Y will likely return the compliment”). In addition to the textual commonsense descriptions, the dataset also contains detailed annotation on the type of causal dimensions, i.e. the events, causes (i.e., `\texttt{xIntent}'), and effects (i.e., `\texttt{xEffect}', `\texttt{oReact}'); with ``\texttt{x}'' and ``\texttt{o}'' pertain to PersonX and others.

We use ATOMIC as our seed to ensure the gold-standard validity of our dataset.
% and we use the train split, as an effort to not leak the ATOMIC test set. 
We synthetically generate \datasetname{} and minimally introduce variance from the LLM by instructing it to be grounded in the context provided by the seed and not to introduce new ones. We mainly utilize GPT-4 series model as the LLM in our data generation pipeline.

\subsubsection{Dataset generation process} 

We prompt LLM to generate the first two premises conditioned on the events, causes (`\texttt{xIntent}', `\texttt{xNeed}', `\texttt{xAttr}'), and effects (`\texttt{xEffect}', `\texttt{xReact}', `\texttt{xWant}', `\texttt{oEffect}', `\texttt{oReact}', `\texttt{oWant}').
% \jh{where are these from? What are they? What does `xIntent', 'xNeed' mean? Are these meta attributes directly from ATOMIC? If so, we should mention it in 4.1. Keep in mind that many readers are not familiar with ATOMIC dataset, and it is really confusing when you just mention these assuming the readers already know}. 
We exclude the static elements, as we want to focus on the dynamic causal relationships where change or action is involved, following the original task~\cite{byrne1989suppressing}. For each event, cause, and effect in ATOMIC, we generate the first two premises in both modus ponens, $p\xrightarrow{}q$ and $p$, and modus tollens, $p\xrightarrow{}q$ and $\neg q$.
% \jh{Because I didn't understand these two things before, I do not understand here}
% , by modifying the second premises. 
Afterwards, we prompt LLM to generate the third premises. We design separately the prompt for the alternative and additional conditions (corresponding to the \textbf{\textcolor{airforceblue}{BM}} and \textbf{\textcolor{amethyst}{BU}} categories) within the context in the first premise. For the alternative condition, we prompt the model to generate conditions that are not related at all to $p$ for the conclusions $q$ to happen. For the additional condition, we prompt the model to generate conditions strongly relate to $p$ for this conclusions $q$ to surely hold. Following the original task setup, we set the same third statement in both cases with modus ponens and modus tollens inferences. 

In our iterations, we discovered that several entities in the ATOMIC dataset are quite abstract, such as ``wants to know what he is selling'' or ``to analyze the thing in question.'' To make these clearer for a general audience and to make them less ambiguous for our study, we prompt LLM to generate more specific examples, changing them to ``asks about the price of a pen'' or ``examine the pen.'' To provide more clarity on the dataset generation process, we attach the samples of prompt and generation process in Appendix \ref{app:prompt_samples_dataset_generation}. Further, to decide the significance of the third premises, whether it serves as alternative or additional condition, we conducted majority voting among multiple human annotators.

% We follow the original suppression task settings and generate datasets for modus ponens and modus tollens inferences. However, we limit ourselves in adopting the experiments on logical fallacies, as we are interested in language model reasoning capabilities, and not on the human fallacies itself. 

% \paragraph{Pragmatic Reasoning: Sarcasm}

% \paragraph{Pragmatic Reasoning: Reference In Context}
% We leverage dataset developed by \citet{ghosh2020interpreting} ... .
% \paragraph{Implicature}

\begin{table}[!t]
\small
\centering
\resizebox{0.95\columnwidth}{!}{%
% \begin{tabularx}{\columnwidth}{p{1.2cm}p{1cm}p{1.8cm}p{1.8cm}}
\begin{tabularx}{\columnwidth}{>{\arraybackslash}m{1.8cm}>{\centering\arraybackslash}m{0.6cm}>{\centering\arraybackslash}m{0.99cm}>{\centering\arraybackslash}m{1.25cm}>{\centering\arraybackslash}m{0.9cm}}
\toprule
\multicolumn{1}{c}{\textbf{Split}} & \textbf{Basic @t} & \textbf{\textcolor{amethyst}{Belief Update}} & \textbf{\textcolor{airforceblue}{Belief Maintain}} & \textbf{All~{w/~3} premises} \\
\midrule
\rowcolor[HTML]{C0C0C0} \multicolumn{5}{l}{\textbf{Inference rule}} \\
\midrule
Modus ponens & 956 & 537 & 335 & 872 \\
Modus tollens & 956 & 537 & 335 & 872 \\
\midrule
\rowcolor[HTML]{C0C0C0} \multicolumn{5}{l}{\textbf{Effect entities}} \\
\midrule
Mental states & 504 & 276 & 184 & 460 \\
Events & 1408 & 798 & 486 & 1284 \\
\midrule
\midrule
\textbf{Total} & \textbf{1912} & \textbf{1074} & \textbf{670} & \textbf{1744} \\
\bottomrule
\end{tabularx}%
}
\caption{Statistics of \datasetname{} dataset.}
\label{tab:dataset_size}
\vspace{-15pt}
\end{table}

\begin{figure*}[!t]
    \centering
    \includegraphics[trim={0, 0.7em, 0, 0}, clip, width=\linewidth]
    {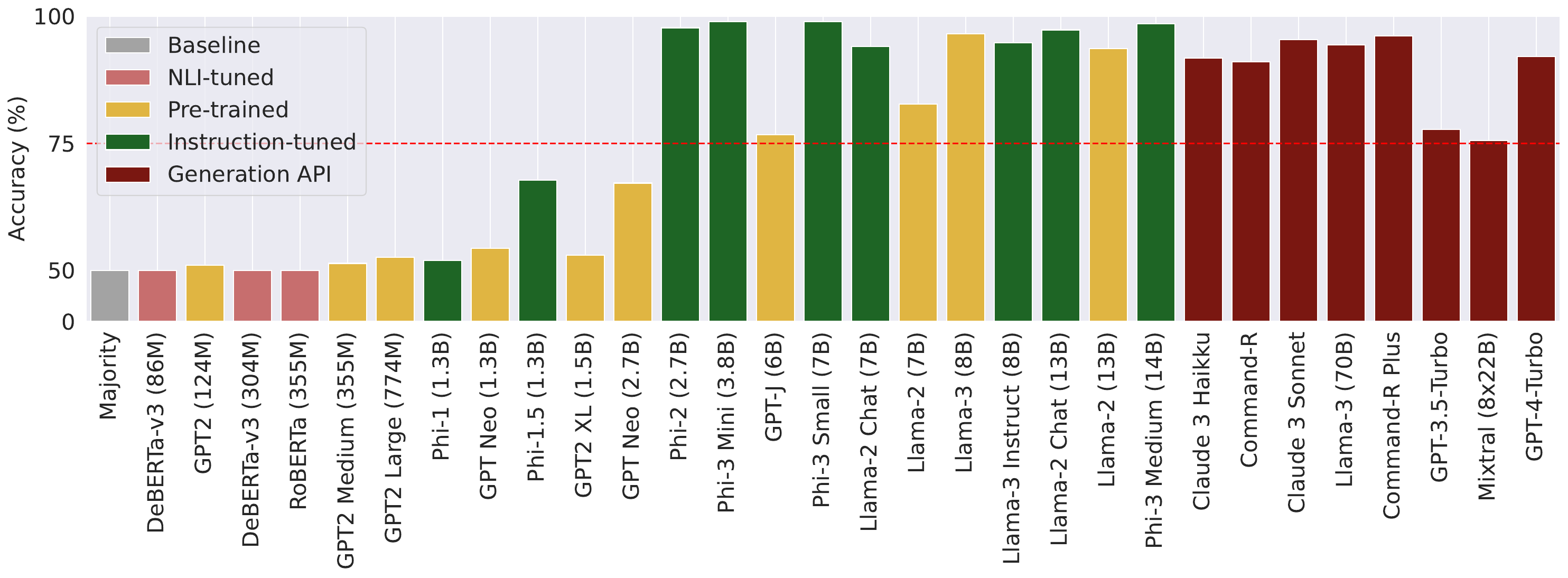}
    \vspace{-10pt}
    \caption{Evaluation on basic logical inference capabilities in \datasetname{} on various LLMs sorted by the \#parameters. Pre-trained LLMs with $\geq$6B parameters achieves adequate accuracy ($\geq$75\%), while instruction-tuned LLMs achieve the same performance on much smaller scale with $\geq$2.7B parameters.}
    \label{fig:two-statement}
    \vspace{-10pt}
\end{figure*}

\subsubsection{Ground-truth formulation}
\label{par:filtering}

To further validate the implied commonsense interaction of the third premises, whether it serves as alternative or additional condition, we manually annotate the final conclusions through a crowdsource annotation task at Appen\cref{foot:appen} (see Appendix~\ref{app:annotation_guidelines}). We cater the variability arises from different interpretations from diverse human readers by asking 5 workers to annotate each problem and then take the majority voting out of them to set the agreed options as the ground truths.
Upon further inspection, we found some annotations that logically invalid, i.e.\ answering $\neg q$ in questions with modus ponens inferences or answering $p$ in modus tollens inferences.
% In such cases, without information about the truth value of $r$, we cannot determine the truth value of Q and thus we cannot conclude $\neg q$. 
We view such cases as non modus ponens (or tollens) inferences and specifically treat the annotation similarly with answering c) $\Diamond q \land \Diamond \neg q$. 

% Since the third statement in both cases with modus ponens and modus tollens inferences are the same, we do the manual annotation and voting on only the samples with modus ponens cases and extend the effect of third premises' significance to the modus tollens cases. We set the ground truth answer of cases with modus tollens inference to be b) $\neg p$ if its modus ponens counterparts get majority vote on a) $q$, declaring that the third premise does not introduce conflict with the prior beliefs. On the other hand, we set the ground truth answer of samples with modus tollens inference to be c) $\Diamond p \land \Diamond \neg p$ when its modus ponens counterparts gets the majority vote on c) $\Diamond q \land \Diamond \neg q$. Here, the third premise is majorly understood as another requirement for the previous conclusion to hold.

In \datasetname{}, both cases of the logical inferences share the same third statement. To streamline our process, we annotate only the modus ponens samples and then extend the insight on the third premises' significance to the modus tollens cases. For modus tollens cases, if the corresponding modus ponens sample primarily supports conclusion a) $q$, indicating no conflict with initial beliefs, we set the correct answer to b) $\neg p$. Conversely, if on the modus ponens samples the majority vote suggests the answer c) $\Diamond q \land \Diamond \neg q$, implying additional requirement for the inference, we likewise categorize the corresponding modus tollens cases answers to be c) $\Diamond p \land \Diamond \neg p$. This process maintains the consistencies of the impact of the third premise effectively across related inference scenarios.

% \vspace{-4pt}

% p -> q<br>
% ~p<br>
% therefore: ~q<br>

% this is a fallacy! -> fallacy, and even we don't mention anything about ~p, so we consider the ground truth answer for them as c, computer should not have human fallacies.
% but ~p is never stated in this case (ponens, suppression, answer c)

\vspace{-3pt}

% \subsection{Context and logical quality checks \\<<not yet>>}
\subsection{Quality check}
% \vspace{-2pt}

\paragraph{Context and logical quality checks} Throughout the data construction phase, we assign one expert to review of the logical formations to ensure they follow the intended structure. We also further gauge the quality of the generated data by reviewing 100 randomly chosen samples to confirm on the context and logical consistency. We conducted a human evaluation via Appen\footnotemark, with three native English speakers assessing each sample's quality. They unanimously confirmed that the conditional relationships in the premises were logically sound across all samples, i.e.\  that $q$ entails $p$ and $q$ entails $r$ in both of the conditional premises. 
% Full details of the evaluation criteria are available in Appendix~\ref{app:quality_check_guideline}. 
We also attach the annotation guidelines in Appendix~\ref{app:annotation_guidelines}.
% Further checks reveals that the synthetic dataset take benefit from the quality of the seed dataset, and our modifications provide minimal variation from the seed dataset. 

% \cite{pan2023logic}

% From the crowd annotators, we also found that 96\% of the samples are following the intended logical structure (using if-then structure). We then take closer a look on the logical structure, and found that the 4\% of the samples is actually have the correct structure. We conjecture that the annotators missed them due to the length of the premises. 
\vspace{-3pt}

\paragraph{Dataset filtering} 

To enhance the quality of our dataset for more reliable evaluation, we refine it by focusing on consensus among annotators. For each question, we utilize answers manually labeled by five independent workers. We measured inter-annotator agreement using Gwet’s AC1~\cite{gwet2008computing}\footnote{\url{https://pypi.org/project/irrCAC/}} since it is better at handling high agreement scenarios, with criteria used in~\cite{wongpakaran2013comparison} to interpret the level of agreement.

We initially measured the annotation agreement which yielded a moderate score of 0.573. Further, we observe that some beliefs are naturally more subjective than others. To make the evaluation dataset more rigorous, we only take the ones with the higher agreement, as commonly practiced in subjective tasks such as sentiment analysis~\cite{bobicev2017inter}. We retain only questions with strong majority agreement (at least 4 out of 5 annotators concurred). Post-filtering, as we retain $\sim$65\% of the original data, the score improved to 0.697 and indicated a substantial agreement. 
% The inter-annotator agreement between workers, using Gwet's AC1~\cite{gwet2008computing}~\footnote{\url{https://pypi.org/project/irrCAC/}}, was substantially high (0.69) on the resultant dataset.

% % From moderate
% AC1 0.48274
% Brennan-Prediger 0.46345
% Fleiss' kappa 0.4202
% Krippendorff's Alpha 0.42025
% Conger's kappa 0.42041
% % To substantial
% AC1 0.69019
% Brennan-Prediger 0.67041
% Fleiss' kappa 0.62217
% Krippendorff's Alpha 0.62221
% Conger's kappa 0.62221

% \begin{figure}[!t]
%     \centering
%     % \includegraphics[width=\linewidth]
%     % {images/two_statement_eval_old.pdf}
%     % \includegraphics[width=\linewidth]
%     % {images/two_statement_eval_full.pdf}
%     \includegraphics[width=\columnwidth]
%     {images/two_statement_lm_eval_filtered.pdf}
%     \caption{Evaluation on simple logical inference capabilities on various models sorted by the \#parameters.}
%     \label{fig:two-statement}
% \end{figure}

\begin{figure*}[!t]
    \centering
    \includegraphics[trim={0, 0.85em, 0, 0}, clip, width=\linewidth]
    {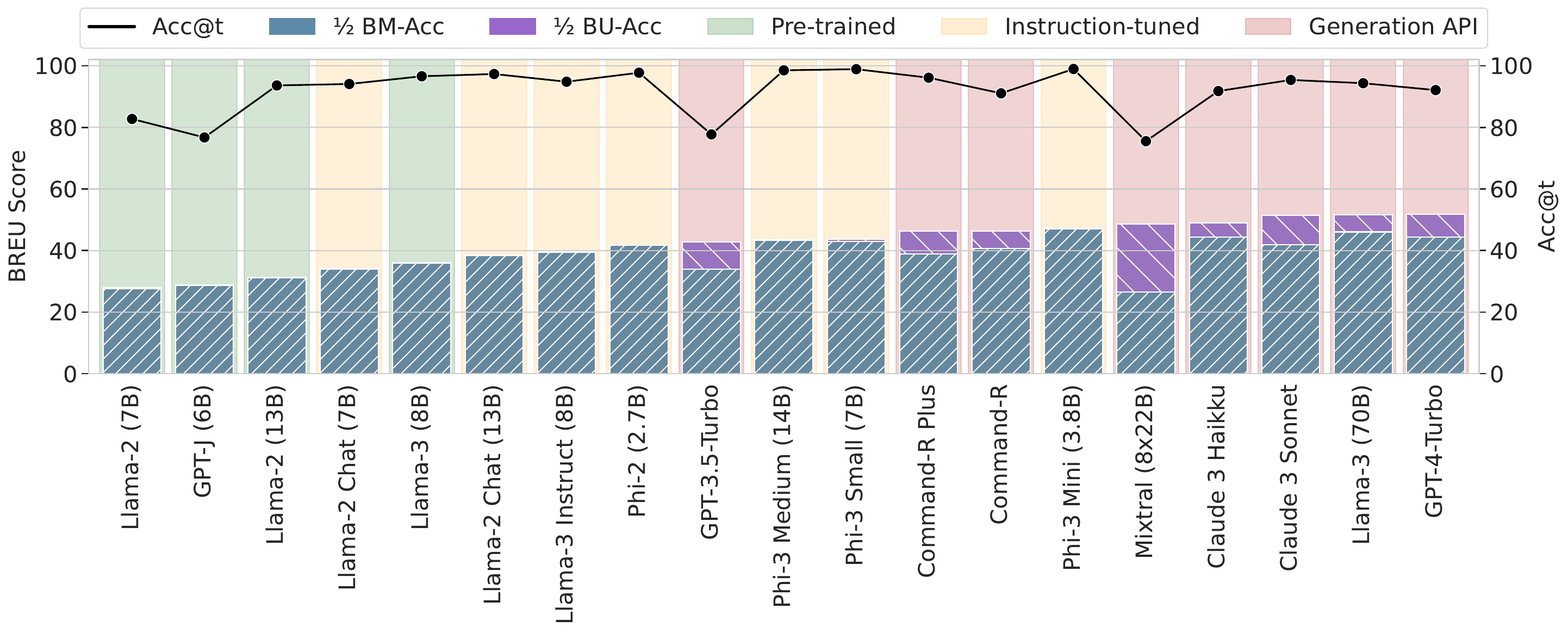}
    \caption{BREU score evaluation on belief revision capabilities in \datasetname{} on various models sorted by the BREU score. While larger-scale LLMs tend to achieve higher BREU score, the performance is far below their basic logical inferences at $t$ (\textbf{Acc@t}), showcasing limited capability of LLMs in performing belief revision.}
    \label{fig:three-statement}
    \vspace{-8pt}
\end{figure*}

\vspace{-3pt}

\subsection{Statistics of \datasetname{}}

% We present the statistics of the final dataset in Table~\ref{tab:dataset_size}. We set the dataset size to an optimal size of approximately 2K entries, considering both representational adequacy and computational limitations associated with large-scale language model inferences. The \textbf{basic @t} category represent the samples with basic logical inference queries at $t$, $\bigchi_t$. The  \textbf{\textcolor{amethyst}{Belief Update}}, \textbf{\textcolor{airforceblue}{Belief Maintain}}, and \textbf{All w/3 premises} category are for th e$\bigchi_{t\Plus1}$ queries at $t\Plus1$. In Table~\ref{tab:dataset_size} also describe the statistics of the inherited categorization of effect entities from ATOMIC, in the case of \texttt{If-Event-Then-Event} and \texttt{If-Event-Then-Mental-State} relations.

Table~\ref{tab:dataset_size} shows the composition of our dataset, sized optimally at around 2K entries to balance representation and computational efficiency for LLM inferences. The dataset includes categories such as \textbf{Basic @t} for basic logical inferences at time $t$, and categories like \textbf{\textcolor{amethyst}{Belief Update}}, \textbf{\textcolor{airforceblue}{Belief Maintain}}, and \textbf{All w/3 premises} for the next step queries at time $t\Plus1$. Additionally, the table details categories inherited from the ATOMIC dataset for the causal relationships of \texttt{If-Event-Then-Event} (e.g., ``promoted to senior manager'') and \texttt{If-Event-Then-Mental-State} (e.g., ``learns something new'').

% with ground truth validated through manual annotations. The resultant dataset is filtered to ensure the quality as an evaluation set

% We first generate and annotate the dataset from the pipeline that intended to output the additional condition for the \textbf{\textcolor{amethyst}{BR}} category. After that, we generate the \textbf{\textcolor{airforceblue}{BM}} part of \datasetname{} using the pipeline intended to output the alternative condition as the third premise.

\footnotetext{\label{foot:appen}\url{https://appen.com/}}

\vspace{-2pt}

\section{Experiment Settings}
\vspace{-4pt}
% \section{Evaluation Settings}

% as the metric to signify model capabilities in our study. Increase of the accuracies in both BM and BU subset (\textbf{BM-Acc} and \textbf{BU-Acc}) indicates better capabilities of belief revision. 

\paragraph{Evaluation metrics} The primary goal of our experiments is to investigate whether LMs possess the capability to perform belief revision in their reasoning processes. Concretely, we consider the models’ predictions on the \textbf{Basic @t} category in the evaluation dataset as the models’ initial belief.  Thus, changes in accuracy on the \textbf{\textcolor{amethyst}{Belief Update}} and \textbf{\textcolor{airforceblue}{Belief Maintain}} categories directly reflects belief revision. We report accuracies in the \textbf{\textcolor{amethyst}{Belief Update (BU-Acc)}} and the
\textbf{\textcolor{airforceblue}{Belief Maintain (BM-Acc)}} subsets to indicate LMs' capabilities in updating and maintaining their beliefs when required. We further introduce a novel metric,
% \textbf{\textcolor{amethyst}{B}\textcolor{airforceblue}{R}\textcolor{amethyst}{E}\textcolor{airforceblue}{U}} (\textcolor{amethyst}{\textbf{B}}elief \textcolor{airforceblue}{\textbf{R}}evision \textcolor{amethyst}{\textbf{E}}valuation \textcolor{airforceblue}{\textbf{U}}nderstudy)
\textbf{BREU} (\textbf{B}elief \textbf{R}evision \textbf{E}valuation \textbf{U}nderstudy), to assess LMs' belief revision ability, by averaging \textbf{\textcolor{amethyst}{BU-Acc}} and \textbf{\textcolor{airforceblue}{BM-Acc}} equally. The goal of BREU
% \textbf{\textcolor{amethyst}{B}\textcolor{airforceblue}{R}\textcolor{amethyst}{E}\textcolor{airforceblue}{U}} 
is to gauge whether the model accurately decides when to update or maintain its prior beliefs. We then benchmark publicly-available LMs and design series of experiments through $\mathbf{\Delta R}$ framework. We perform zero-shot-classification on series of smaller to larger scales pre-trained and finetuned LMs, and prompt LLMs generations through API.

\vspace{-4pt}

\begin{figure*}[!t]
  \centering
  \begin{subfigure}[t]{0.32\linewidth}
  \hspace*{-15pt}
      \includegraphics[trim={0, 0, 0, 0}, clip, width=\linewidth]{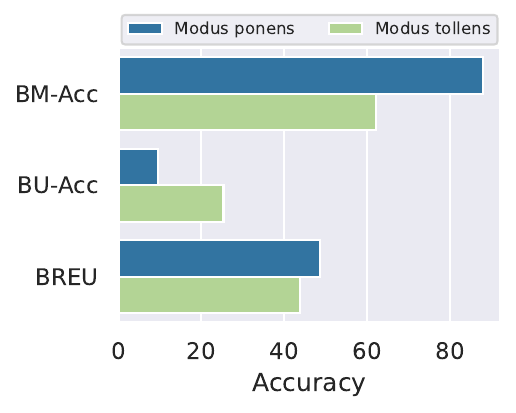}
      \vspace{-5pt}
      \caption{Modus Ponens / Modus Tollens}
      \label{fig:ponens-tolens}
  \end{subfigure}
  \begin{subfigure}[t]{0.32\linewidth}
    \centering
    \hspace*{2pt}
      \includegraphics[trim={0, 0, 0, 0}, clip, width=0.83\linewidth]{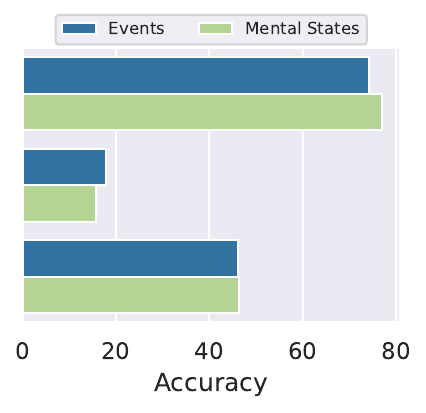}
      \vspace{-5pt}
      \caption{Events / Mental States}
      \label{fig:event_mentalstate}
  \end{subfigure}
  \begin{subfigure}[t]{0.32\linewidth}
    \centering
    \hspace*{2pt}
      \includegraphics[trim={0, 0, 0, 0}, clip, width=0.818\linewidth]{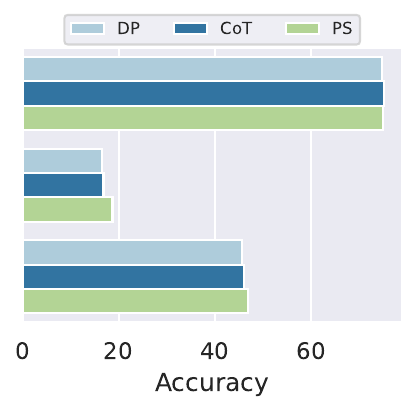}
      \vspace{-5pt}
      \caption{Prompting Methods}
      \label{fig:prompt_modes}
  \end{subfigure}
  \caption{Performance comparisons dissected across various aspects covering distinction on modus ponens and modus tollens, on different effect entities, and on different prompt methods.}
  \label{fig:analysis_discussion}
  \vspace{-10pt}
\end{figure*}

% We perform zero-shot classification using GPT-2~\cite{radford2019language}, Llama-2~\cite{touvron2023Llama}, Llama-3~\cite{Llama3modelcard}, Phi~\cite{gunasekar2023textbooks, li2023textbooks}, GPT Neo~\cite{gao2020pile}, and GPT-J~\cite{gpt-j}, and using NLI finetuned models: RoBERTa~\cite{liu2019roberta}, DeBERTa-v3 base~\cite{laurer2024less}, and DeBERTa-v3 large~\cite{laurer_building_2023}, that has been trained on various NLI datasets.
\paragraph{Models} We perform zero-shot classification using 
% two types of LMs: 
% both 
encoder-only and decoder-only LMs. For encoder-only LMs, we employ entailment-based inference~\cite{yin-etal-2019-benchmarking} using NLI-finetuned LMs of RoBERTa~\cite{liu2019roberta}, DeBERTa-v3 base~\cite{laurer2024less}, and DeBERTa-v3 large~\cite{laurer_building_2023}. For decoder-only LMs, we follow \citet{brown2020gpt3}
% likelihood-based zero-shot inference~\cite{brown2020gpt3} 
using GPT~\cite{radford2019language,gpt-neo,gpt-j}, Llama~\cite{touvron2023Llama,touvron2023Llama2,Llama3modelcard}, and Phi series~\cite{gunasekar2023textbooks,li2023textbooks,abdin2024phi3}.
% , and using NLI finetuned models: RoBERTa~\cite{liu2019roberta}, DeBERTa-v3 base~\cite{laurer2024less}, and DeBERTa-v3 large~\cite{laurer_building_2023}, that has been trained on various NLI datasets.
% Provided that all larger-scale LMs up to 15B parameters fail on belief revision, 

We 
% further expand our evaluation to even 
also include larger-scale LLMs with $\geq$35B parameters. We evaluate the belief revision capability of these larger-scale LLMs via completion API through generation-based approach. 
We employ three zero-shot prompting methods, i.e., \textbf{direct prompting (DP)}, triggering the generation of \textbf{chain-of-thought (CoT)}~\cite{kojima2022large}, or through \textbf{plan and solve (PS)} prompting~\cite{wang-etal-2023-plan}. We employ 8 large-scale LLMs, i.e., Llama-3 70B~\cite{Llama3modelcard}, Mixtral 8x22B~\cite{jiang2024mixtral}, Command R, Command R+~\cite{command_r_Cohere_2024}, Claude 3 Haiku, Sonnet~\cite{claude_3_Anthropic_2024}, GPT-3.5 Turbo, and GPT-4 Turbo~\cite{openai2023gpt4}. Here, we follow~\citet{yao2022react} and instruct the model to output the exact character of the final answer as a format. We then retrieve the final answer and report accuracy of the final answer as the metric. When the answer does not follow the format instructed before, we treat it as an instruction-following error.

% to perform a comparative assessment of the hypotheses’ plausibility regarding the changing or conflicting pieces of information.

% We need to show with observation on this dataset, that it’s just not that it’s difficult to answer the QA in the suppression test format, but it’s the non-monotonic reasoning part that makes it difficult

% \section{Result and Analysis}

% \paragraph{LLMs are able to generalize on simple two statements inference}

% \paragraph{When should we care about belief revision?}
% % We also report the majority class baseline (∼33\% since our labels are balanced).
% Belief revision is a crucial aspect of reasoning. Nonetheless, this ability needs to be taken into consideration only if the models have adequate on simple logical reasoning. To assess the adequacy of simple logical reasoning, we evaluate a wide range of models in performing modus ponens and modus tollens using the first two statements in \datasetname{}. As shown in Figure~\ref{fig:two-statement}, scaling the number of parameters correlate with the model capability on simple logic. Smaller pre-trained LMs with <2B parameters are all unable to perform basic logic task achieving almost random performance close to the majority baseline. pre-trained LLMs with 6B parameters or above achieve >75\% accuracy, while instruction-tuned LLMs show an emerging ability from 2.7B parameters achieving a significantly higher performance with >90\% accuracy.

\vspace{-1pt}

\section{Result and Analysis}

\paragraph{Smaller models fail even on basic logical reasoning tasks.}

% Following the $\mathbf{\Delta R}$ framework, 
We start by examining the inferences through the first two premises in \datasetname{}. In Figure~\ref{fig:two-statement}, we find as the number of parameters in LMs increases, their ability to handle basic logical inference improves. Smaller models, with $<$2B parameters, struggle with these tasks, scoring close to the majority baseline. Models $>$6B parameters do better, surpassing 75\% accuracy. Pre-trained LMs with ${>}$6B parameters achieve $\geq75\%$ accuracy, while instruction-tuned LMs show an emerging ability from 2.7B parameters achieving significantly higher performance with $\geq90\%$ accuracy.

\paragraph{LLMs are incapable of revising their prior beliefs.} 
We group our further exploration on these LMs that performed well ($\geq75\%$ accuracy) in basic logical inference, and evaluate their average performance in Belief Maintain (BM) and Belief Update (BU) subsets. Despite being a strong reasoner on simple logic, all larger-scale LMs under study fail to perform well on these subsets of \datasetname{}. In evaluation shown in Figure~\ref{fig:three-statement}, most of the non-API based models perform almost $0\%$ in BU-Acc, indicating their inability on performing belief revision. We observe that all larger-scale both open-source and commercial LLMs perform better on the belief revision tasks, but their performances are still very limited, achieving at most $\sim$50\% on BREU.

% Provided that all the previously well-performing larger-scale LMs fail on belief revision, we further expand our evaluation through generation-based approach utilizing even larger LLMs and note the findings in Table~\ref{tab:three-statement-llm}. 

% The data indicates most significant improvements in the BU subset, though overall belief revision improvements remain marginal, showing $\sim$1\% increase in BREU.

\paragraph{LLMs confront a trade-off between updating and maintaining their prior beliefs.}

We discover a trade-off between BU-Acc and BM-Acc: models performing well on one subset typically faltered on the other, especially in models where the BU-Acc is not close to $0\%$ (see Fig~\ref{fig:three-statement}). This indicates a potential tension between enhancing specific capabilities, as improving one aspect could inadvertently weaken another. An ideal model would excel at belief revision by consistently making the right decision on whether the new information conflicts with prior beliefs or aligns with them.
% updating or maintaining prior beliefs.
This underlines the importance of developing strategies that refine the ability to revise beliefs accurately, ensuring its reliability across various scenarios.

% On the series of models that are declared superior than its counterparts, i.e.\ GPT-3.5 and GPT-4, we observe that their average accuracy is increasing, except for Command R and Command R+ that having slight performance drop. 

% More importantly, most LLMs achieve only $10{-}20\%$ accuracy on samples that require update of belief (BU-Acc), with the exception on Mixtral (8x22B) model that achieves 35.38\% BU-Acc score. Nevertheless, this model also comes with the lowest accuracy on samples that do not require revision (BM-Acc) and only achieves 36.57\% in BM-Acc, while others achieve ${>}50\%$ scores.

\section{Discussion}

\paragraph{Belief revision is harder in a more complex task with modus tollens inferences.}

% \begin{figure}[ht]
%     \centering
%     \includegraphics[width=\columnwidth]
%     {images/ponens_tollens.pdf}
%     \vspace{-20pt}
%     \caption{Performances in tasks with modus ponens and modus tollens inference rule. Tasks with modus tollens rule show reduced accuracy, as they are more complex.}
%     \label{fig:ponens_tollens}
% \end{figure}
% Despite higher BU-Acc in modus tollens, its BM-Acc and BREU scores are lower compared to modus ponens tasks.

We compare LLMs' belief revision capabilities in average, through tasks with modus ponens and modus tollens rule as the basic logical inferences at step $t$. As observed in Figure~\ref{fig:ponens-tolens}, LLMs show reduced BREU score in tasks with modus tollens rule. This is expected, as modus tollens is inherently more difficult relative to modus ponens as it require backward directions of reasoning and it involves reasoning with negations~\cite{evans1982psychology, evans1993mental, girotto1997effect}. Furthermore, in tasks involving modus tollens inference, we observe a notably higher BU-Acc compared to a much lower BM-Acc. This disparity suggests that executing accurate belief revision becomes more challenging in complex tasks: decisions to update or maintain beliefs are less clear-cut in these scenarios compared to simpler tasks.

\paragraph{Belief update on abstract concept is more challenging for LLMs.}

% \paragraph{Belief update on mental states effect entities is more challenging than events.}

% \begin{figure}[ht]
%     \centering
%     \includegraphics[width=\columnwidth]
%     {images/event_mentalstate.pdf}
%     \vspace{-20pt}
%     \caption{Performance of models on tasks involving different effect entities. Models find belief update about mental state causes harder than about event causes.}
%     \label{fig:event_mentalstate}
% \end{figure}

We examine LLMs' belief revision capabilities in average, when dealing with scenarios involving causal relationships on events and mental states effect entities and note them in Figure~\ref{fig:event_mentalstate}. While the BREU score is similar, LLMs demonstrate tendency towards maintaining their beliefs in mental state effects instead of updating them. This may stem from the challenge of recognizing additional requirements implied from the third, mental state-related, premise which is inherently more abstract and less directly observable than a concrete sequential event.

\paragraph{Better prompting method does not help on belief revision.}
% \paragraph{Better prompting methods yield limited gain on belief revision.}

% \begin{figure}[ht]
%     \centering
%     \includegraphics[width=\columnwidth]
%     {images/prompt_modes.pdf}
%     \vspace{-20pt}
%     \caption{Performances from different prompting methods. Prompting methods such as CoT and PS, only yield a marginal improvement ($\sim$1\%) compared to DP.}
%     \label{fig:prompt_modes}
% \end{figure}

We explore how different prompting methods affect belief revision abilities of LLMs on average. Figure \ref{fig:prompt_modes} shows that CoT, which encourages LLMs to elicit reasoning steps, does not significantly enhance belief revision. While this may stem from its vulnerability to missing-step errors~\cite{wang-etal-2023-plan}, attempts to correct these errors with the PS prompting offer minimal benefits, improving only by $\sim$1\% of BREU. This suggests the ability to revise beliefs could still be absent despite elicitation of reasoning steps. We put a more detailed analysis in  Appendix~\ref{app:variation_prompt_impact}.

\section{Conclusion}

The ability to reason and adapt to changing information is crucial for NLP applications in the real world. Most evaluations assume static knowledge environment, which does not prepare models for dynamic real-life scenarios. To address this, \datasetname{} is introduced as a diagnostic dataset for evaluating belief revision capability in LMs. Through \datasetname{} and a novel evaluation framework for evaluating reasoning in a dynamically evolving environment, $\mathbf{\Delta R}$, we reveal that current models struggle with updating their beliefs in response to new information, highlighting the need for improved adaptability and reliability. By exposing these limitations, our work underscores the imperative of developing AI models capable of reasoning adeptly with evolving data, thereby propelling them closer to real-world applicability and robustness.

\section{Future Work}

Observing the limited gain offered by the current prompting strategy, we suggest that future developments aim to enhance LMs' understanding and handling of premise dependencies. A particularly promising direction is to integrate LMs with deterministic symbolic solvers to convert problems into logical rules through chain-of-thought reasoning to aim for a deeper comprehension of the intricate dependencies among different premises. Future work could aim to overcome current limitations and pave the way for AI systems to be reliable across evolving scenarios.

\section*{Acknowledgements}

We thank Yejin Bang for the insightful discussions. This work has been partially funded by the PF20-43679 Hong Kong PhD Fellowship Scheme and the Hong Kong Fellowship Scheme by the Hong Kong Research Grants Council (RGC), and Care-E Project (FS116).

\section*{Limitations}

\paragraph{Towards understanding general belief revision capabilities.} 
% further, about internal beliefs
% We are limited in following one form of belief revision test, i.e.\ with conditionals. While there are many 
% Here it involves belief derived by modus ponens modus tollens and the need of commonsense reasoning
Our study on belief revision using the \datasetname{} dataset via the $\mathbf{\Delta R}$ framework focuses on belief changes driven by logical inferences like modus ponens and modus tollens, which may not fully represent the complexity of real-world belief revision that often includes a broader range of scenarios and subtleties. To add, our methodology primarily considers the introduction of new premises as the trigger for belief revision, overlooking how beliefs might change through re-evaluation of existing knowledge or shifts in perspective in the absence of new information (i.e.\ in~\citet{kronemyer2014non}). 

Furthermore, how to define models’ beliefs is a debatable question and there are some works inspecting models’ internal beliefs by probing the hidden states~\cite{burnsdiscovering, zou2023representation}. n our work, however, we conceptualize "belief" similarly to its usage in dialogue systems, where it represents what the system currently considers true based on the context~\cite{feng2023towards, van2020knowing}. Further exploration on examining the belief through other approaches (e.g., probing hidden states of LLMs) lies outside this paper’s scope and we leave it as future work.

% Lastly, our approach to simulate future data is constrained by our inability to determine what LMs have previously known and by resource limitations that restrict the training of large-scale models from scratch.

\paragraph{Intersection of reasoning capability and knowledge capacity.} 
The evaluation of models' reasoning capabilities is intricately tied to their knowledge capacity, presenting a significant challenge in discerning pure reasoning capability from mere knowledge recall. Current benchmarks often fail to disentangle these aspects, as models with extensive knowledge bases may appear to possess superior reasoning abilities when, in fact, they might be leveraging stored information rather than demonstrating genuine inferential logic. This conflation complicates the assessment of a model's true reasoning faculties, as performance improvements on reasoning tasks could be attributed to enhanced information retrieval rather than advancements in reasoning algorithms. Similar to observations in other reasoning datasets, we acknowledge the limitation that the improved performance of models tested on \datasetname{} might not only stem from their ability to revise beliefs but could also be influenced by superior knowledge recall~\cite{huang-chang-2023-towards}. Future research could delve deeper into the relationship between these capabilities, specifically focusing on developing evaluation methods that effectively distinguish between them.

\section*{Ethics statement}

This research explores how well LMs can revise their beliefs when faced with new information, which is crucial for their use in constantly changing real-world situations. We created a reasoning evaluation dataset to test whether LMs can revise their beliefs correctly or if they stick to their initial assumptions. This is important for using LMs in areas where being accurate and up-to-date is vital, like healthcare or legal advice. In example, being able to revise beliefs appropriately could help prevent LMs from repeating outdated or wrong information, making them more reliable and trustworthy. Plus, LMs that can refresh their understanding according to new societal norms can avoid perpetuating biases, contributing to the fair and ethical use of AI. We consider this a promising and significant area for research. We construct the dataset using events, causes, and effects from ATOMIC and the construction template is designed and reviewed manually and attached in this paper. We utilized crowd-sourced annotators who voluntarily participated through the platform Appen\cref{foot:appen}, choosing tasks they deemed fairly compensated. The annotators were presented with multiple-choice tasks predefined to avoid bias and protect privacy, ensuring an ethical annotation process.

\bibliography{anthology,custom}

\newpage
\appendix
% \appendixpage

\setcounter{table}{0}
\renewcommand{\thetable}{A\arabic{table}}
\setcounter{figure}{0}
\renewcommand{\thefigure}{A\arabic{figure}}

\clearpage
\section*{Appendix}

% We further provide appendixes that consists the following contents:
% \begin{itemize}
%     \item[\ref{app:explain_arg_strength}.] Detailed Explanation on Additional and Alternative Arguments
%     \item[\ref{app:prompt_samples_dataset_generation}.] Samples of Prompts 
%     % Samples and Dataset Generation Process
%     \item[\ref{app:annotation_guidelines}.] Annotation Guidelines
%     \item[\ref{app:generation_samples}.] Sample of Model Outputs
%     \item[\ref{app:additional_analysis}.] Additional Analysis
% \end{itemize}

% \section{Detailed Explanation on Additional and Alternative Arguments}
% \label{app:explain_arg_strength}
% Explain more why n or u is strong or weak

\section{Samples of Prompts}
% \section{Prompt Samples and Dataset Generation Process}
\label{app:prompt_samples_dataset_generation}

\begin{figure*}[!t]
    \centering
    \includegraphics[width=\linewidth]
    {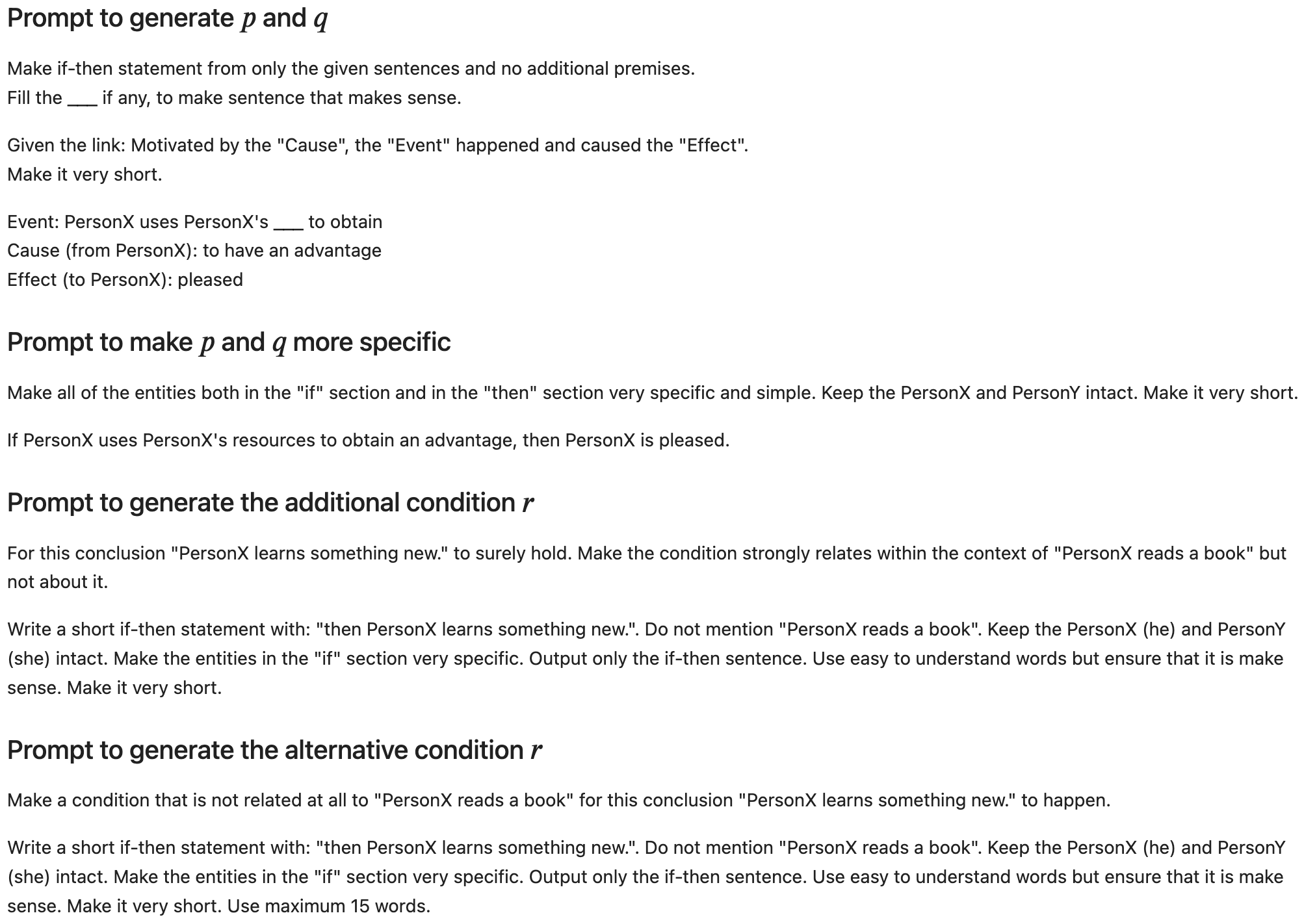}
    \vspace{-20pt}
    \caption{
    Samples of prompts utilized in each of the \datasetname{} generation pipeline. Here, we take the \texttt{Event: PersonX uses PersonX's \_\_\_ to obtain}, \texttt{Cause (from PersonX): to have an advantage}, \texttt{Effect (to PersonX): pleased} from ATOMIC, to generate $p$, $q$, and $r$ for us to form queries at step $t$ and $t\Plus1$ in \datasetname{} and later go through the manual annotaion process.
    }
    \label{fig:prompt_samples}
\end{figure*}

To provide more clarity on the dataset generation process, we attach the samples of prompt in Figure~\ref{fig:prompt_samples}.

\section{Annotation guidelines}
\label{app:annotation_guidelines}

\begin{figure*}[!t]
\centering
\begin{subfigure}{\linewidth}
\centering
  \includegraphics[width=\linewidth]{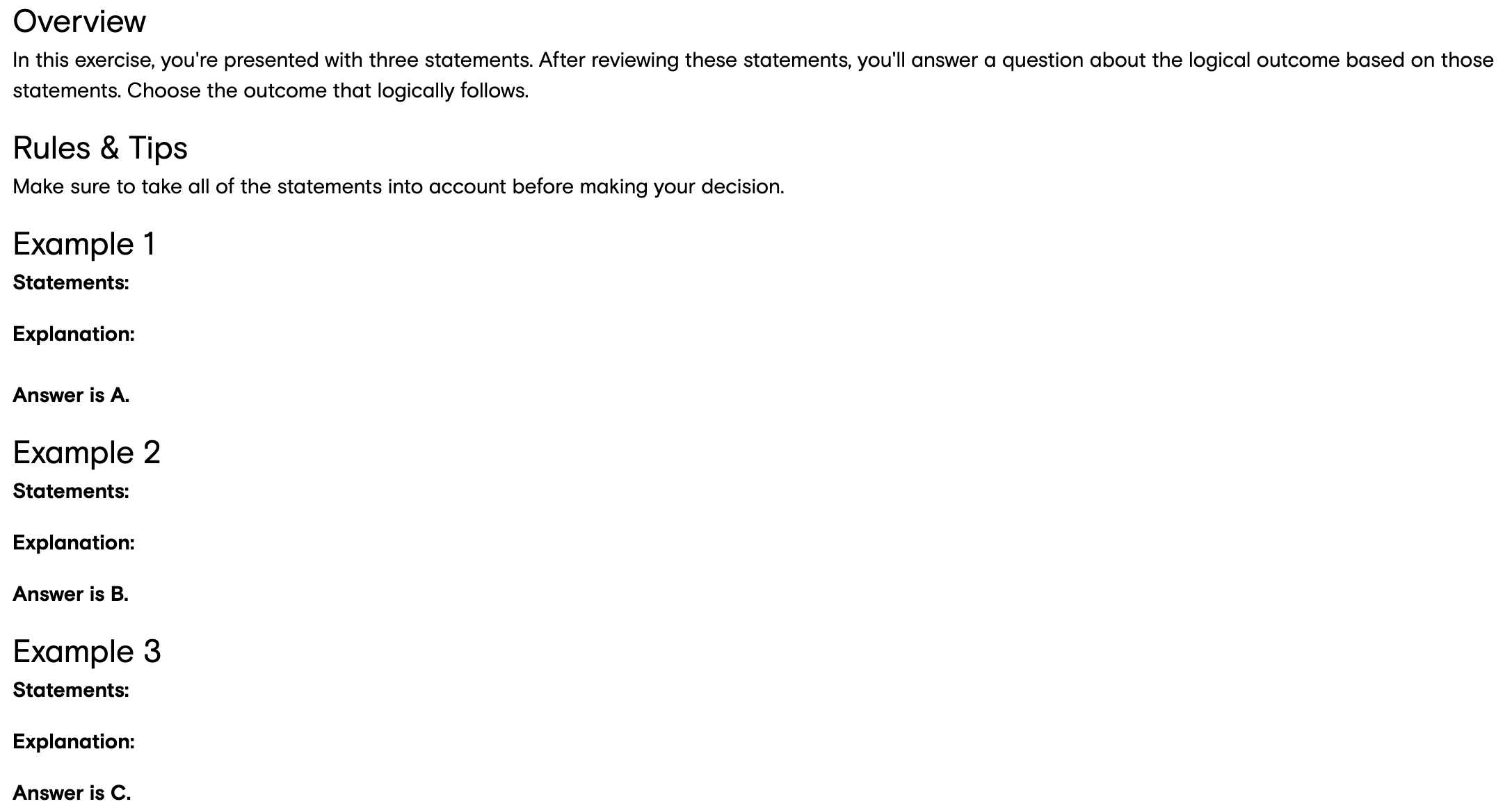}
  \caption{Annotation Guidelines}
  \label{fig:GT_annotation_guideline}
\end{subfigure}
\begin{subfigure}{\linewidth}
\centering
  \includegraphics[width=\linewidth]{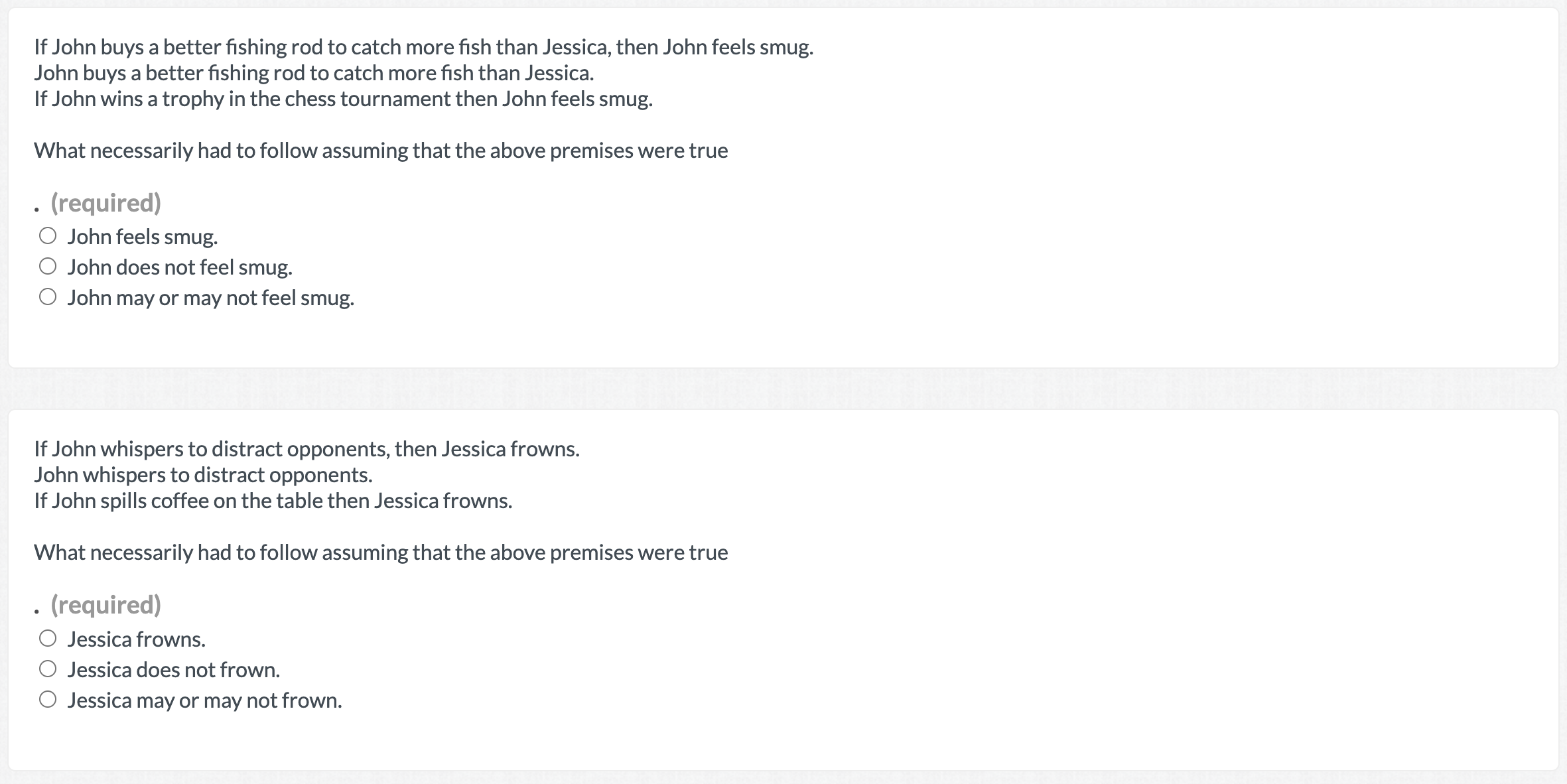}
  \caption{Example of Annotation Questions}
  \label{fig:GT_annotation_example}
\end{subfigure}
\caption{Details on ground truth annotation
}
\label{fig:GT_annotation}
\end{figure*}

\begin{figure*}[!t]
\centering
\begin{subfigure}{\linewidth}
\centering
  \includegraphics[width=\linewidth]{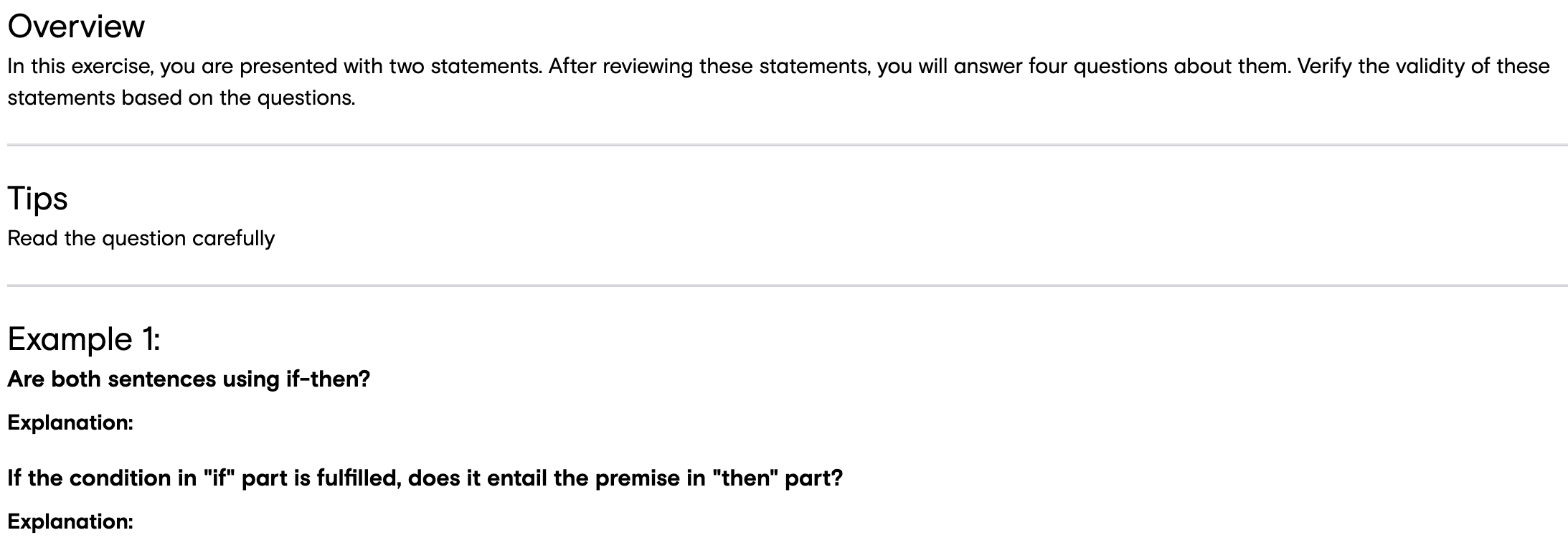}
  \caption{Annotation Guidelines}
  \label{fig:Eval_annotation_guideline}
\end{subfigure}
\begin{subfigure}{\linewidth}
\centering
  \includegraphics[width=\linewidth]{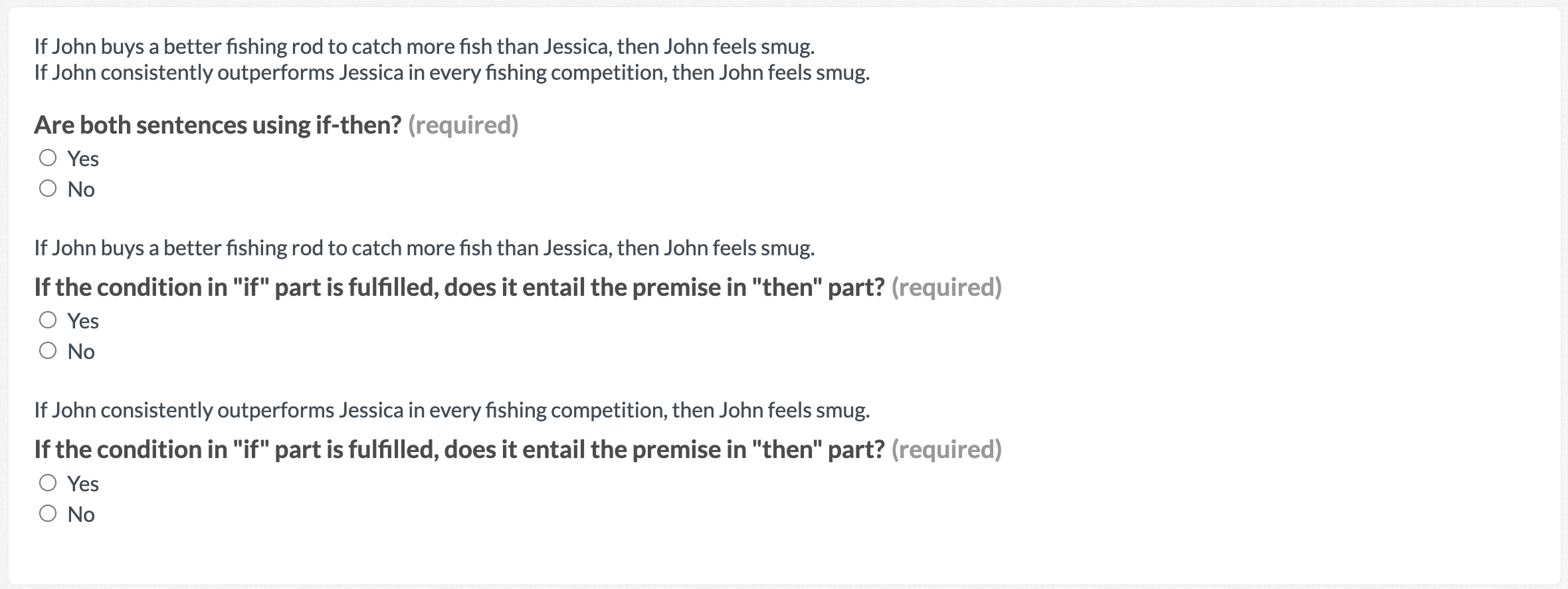}
  \caption{Example of Annotation Questions}
  \label{fig:Eval_annotation_example}
\end{subfigure}
\caption{Details on quality check annotation
}
\label{fig:Eval_annotation}
\end{figure*}

We provide human annotators with specific guidelines and examples, as detailed in Figures \ref{fig:GT_annotation} and \ref{fig:Eval_annotation} for ground truth and quality check annotations, respectively.

% \section{Sample of Model Outputs}
% \label{app:generation_samples}

\section{Additional analysis}
\label{app:additional_analysis}

\subsection{LLMs logical reasoning ability are not robust in the presence of distractors} 

\begin{figure*}[!t]
    \centering
    \includegraphics[width=\linewidth]
    {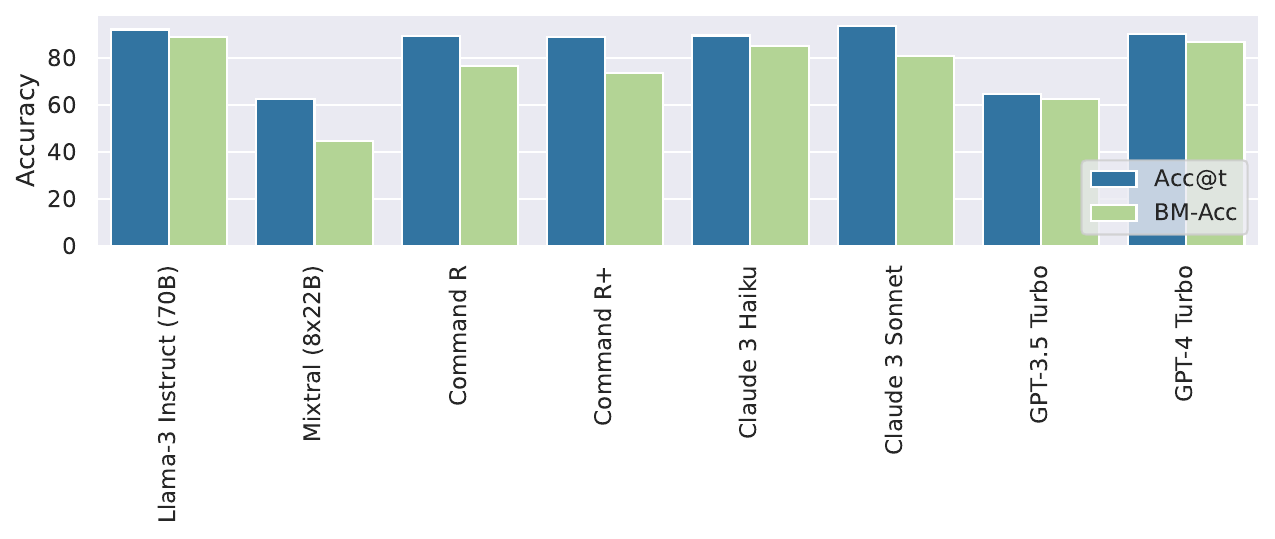}
    \vspace{-20pt}
    \caption{
    LLMs show decreased inference performance when exposed to noise from the new information in alternative condition.
    }
    \label{fig:distractors}
\end{figure*}

We analyze the performance of LLMs on basic logical inference tasks and compare it to their accuracy on the BM subset, which differs only by including a third premise. We selected 378 queries from the \datasetname{} dataset where premises overlap between the basic logical inference tasks at time $t$ and the BM subset for a fair comparison, and visualize them in Figure~\ref{fig:distractors}. On most of the models, LMs' accuracy on samples that do not require change of conclusion (BM-Acc) is dropping compared to its basic inference at $t$ performances. This indicates that the logical reasoning ability of these models are not robust in the presence of distractors, exposing a critical problem of these models especially on the challenges in currently adopted retrieval-augmented-generation (RAG) pipeline to manage noisy documents that have question-related content despite lacking substantive information~\cite{lewis2020retrieval, chen-etal-2022-murag, gao2023retrieval}.

\subsection{Details on prompting methods variations on their gain at belief revision.}
\label{app:variation_prompt_impact}

\begin{table}[!t]
    \small
    \centering
    \resizebox{\columnwidth}{!}{
    % \begin{tabular}{l|l|c|c|c}
    \begin{tabularx}{\columnwidth}{>{\arraybackslash}p{1.3cm}|>{\arraybackslash}p{0.9cm}|>{\centering\arraybackslash}p{1.1cm}|>{\centering\arraybackslash}p{1.2cm}|>{\centering\arraybackslash}p{1.1cm}}
    \toprule
        \multicolumn{1}{l|}{\textbf{Models}} & \multicolumn{1}{l|}{\textbf{Method}} & \textbf{BU-Acc} & \textbf{BM-Acc} & \textbf{BREU} \\
    \midrule
        \multirow{3}{\hsize}{Llama-3 Instruct (70B)} & DP & 10.99\% & 92.09\% & 51.54\% \\
         & CoT & 12.57\% & 89.40\% & 50.99\% \\
         & PS & 12.66\% & 88.21\% & 50.44\% \\
         \midrule
        \multirow{3}{\hsize}{Mixtral (8x22B)} & DP & 35.38\% & 36.57\% & 35.98\% \\
         & CoT & 27.28\% & 34.93\% & 31.11\% \\
         & PS & 44.04\% & 53.13\% & 48.59\% \\
         \midrule
        \multirow{3}{\hsize}{Command R} & DP & 12.10\% & 80.45\% & 46.28\% \\
         & CoT & 11.36\% & 81.19\% & 46.28\% \\
         & PS & 19.37\% & 69.85\% & 44.61\% \\
         \midrule
        \multirow{3}{\hsize}{Command R+} & DP & 13.69\% & 75.67\% & 44.68\% \\
         & CoT & 14.71\% & 77.76\% & 46.24\% \\
         & PS & 13.41\% & 65.07\% & 39.24\% \\
         \midrule
        \multirow{3}{\hsize}{Claude-3 Haiku} & DP & 9.40\% & 88.66\% & 49.03\% \\
         & CoT & 13.50\% & 83.73\% & 48.62\% \\
         & PS & 13.22\% & 82.99\% & 48.11\% \\
         \midrule
        \multirow{3}{\hsize}{Claude-3 Sonnet} & DP & 19.65\% & 82.69\% & 51.17\% \\
         & CoT & 21.51\% & 81.19\% & 51.35\% \\
         & PS & 16.76\% & 83.73\% & 50.25\% \\
         \midrule
        \multirow{3}{\hsize}{GPT-3.5 Turbo} & DP & 14.53\% & 55.22\% & 34.88\% \\
         & CoT & 20.48\% & 65.22\% & 42.85\% \\
         & PS & 17.78\% & 67.91\% & 42.85\% \\
         \midrule
        \multirow{3}{\hsize}{GPT-4 Turbo} & DP & 16.76\% & 86.72\% & 51.74\% \\
         & CoT & 13.59\% & 87.76\% & 50.68\% \\
         & PS & 12.76\% & 88.66\% & 50.71\% \\
    \bottomrule
    \end{tabularx}
    }
    \caption{The effectiveness of various prompting techniques varies across LLMs and subset of \datasetname{}, enhancing performance in some while degrading it in others.}
    \label{tab:three-statement-llm}
\end{table}

We provide more details on the investigation in the impact of varied prompting techniques on the performance accuracy of several models, as summarized previously in Figure~\ref{fig:prompt_modes}. In that figure, the data indicates most significant performance improvements in the BU subset, though overall belief revision improvements remain marginal, showing $\sim$1\% increase in BREU.  In examining the performance across models and different prompting methods as shown in Table~\ref{tab:three-statement-llm}, it is clear that the influence of these methods is not uniform. For instance, the PS prompting method notably boosted accuracy for models like Mixtral 8x22B and Command R by over 10\%. Conversely, this same strategy led to performance reductions in models such as Claude-3 Sonnet and GPT-4 Turbo. Similarly, utilizing CoT and PS exhibited mixed outcomes across models. It strengthened robustness in models like GPT-3.5 Turbo and GPT-4 Turbo, as shown by higher BM-Acc scores, while it increased sensitivity to noise in models like Llama-3 Instruct (70B) and Command R, resulting in reduced BM-Acc values.

\end{document}